\documentclass[12pt,headings=standardclasses]{scrartcl}
\usepackage{amsmath}
\usepackage{amsthm}
\usepackage{mathptmx}
\usepackage{newtxmath}

\usepackage[T1]{fontenc}
\usepackage[utf8]{inputenc}
\usepackage{geometry}
\usepackage{color}
\usepackage{array}
\usepackage{verbatim}
\usepackage{float}
\usepackage{rotfloat}
\usepackage{booktabs}
\usepackage{bm}
\usepackage{graphicx}
\usepackage{setspace}
\makeatletter

\providecommand{\tabularnewline}{\\}

\theoremstyle{plain}
\newtheorem{thm}{\protect\theoremname}[section]
\theoremstyle{plain}
\newtheorem{prop}[thm]{\protect\propositionname}

\@ifundefined{date}{}{\date{}}

\usepackage{amsthm}
\theoremstyle{definition}

\usepackage{cleveref}
\usepackage{algorithmic}
\usepackage{array}
\usepackage{stfloats}
\usepackage[super]{nth}
\usepackage{subcaption}
\usepackage{caption}
\usepackage{xcolor}
\usepackage{tikz}
\usepackage{pgf}

\g@addto@macro\@floatboxreset\centering

 \AtBeginDocument{
 	\let\ref\Cref
 }

\crefalias{prop}{proposition}

\usepackage{tikz}

\usetikzlibrary{arrows}

\usetikzlibrary{bayesnet}
\usetikzlibrary{decorations.pathreplacing}

\tikzset{
	diagonal fill/.style 2 args={fill=#2, path picture={
			\fill[#1, sharp corners] (path picture bounding box.south west) -|
			(path picture bounding box.north east) -- cycle;}},
	reversed diagonal fill/.style 2 args={fill=#2, path picture={
			\fill[#1, sharp corners] (path picture bounding box.north west) |- 
			(path picture bounding box.south east) -- cycle;}}
}

\tikzstyle{partialobs} = [latent,diagonal fill={gray!25}{gray!0}]

\usepackage[style=authoryear,citestyle=authoryear,isbn=false,url=false,eprint=false,minbibnames=10, maxbibnames=10,maxcitenames=3,firstinits=true]{biblatex}
\DeclareFieldFormat[article,inbook,book,incollection,inproceedings,patent,thesis,unpublished]{citetitle}{#1}
\DeclareFieldFormat[article,inbook,incollection,inproceedings,patent,thesis,unpublished]{title}{#1} 

\providecommand{\propositionname}{Proposition}
\providecommand{\theoremname}{Theorem}

\begin{document}
\global\long\def\Tsq{T^{2}}%
\global\long\def\priordist{p_{z}}%
\global\long\def\detencoding{g_{\mbphi}}%
\global\long\def\detdecoding{f_{\mbtheta}}%
\global\long\def\encoding{q_{\mbphi}(\mbz\g\mbx)}%
\global\long\def\decoding{p_{\mbtheta}(\mbx\g\mbz)}%
\global\long\def\czero{c_{0}}%
\global\long\def\dataset{\mathcal{D}}%
\global\long\def\mbxover#1{\mbx^{(#1)}}%
\global\long\def\stdGauss{\Norm(\mbzero, \mbI)}%
\global\long\def\pdata{p_{data}(x)}%
\global\long\def\ptheta{p_{\mbtheta}}%
\global\long\def\pthetaxgivenz#1#2{\ptheta(#1 \g#2)}%
\global\long\def\qphi{q_{\mbphi}}%
\global\long\def\qphizgivenx#1#2{\qphi(#1 \g#2)}%
\global\long\def\pz{p(\mbz)}%
\global\long\def\oursacronym{SPE}%
\global\long\def\TsqKLD{T_{\mathrm{KLD}}^{2}}%
\global\long\def\QERE{Q_{\mathrm{ERE}}}%
\global\long\def\etal{\textit{et al}. }%
\global\long\def\ie{\textit{i}.\textit{e}.}%
\global\long\def\eg{\textit{e}.\textit{g}.}%

\global\long\def\g{\,|\,}%
\global\long\def\gg{\,\|\,}%
\global\long\def\KL#1#2{\textrm{KL}\left(#1\gg#2\right)}%
\global\long\def\E{\mathbb{E}}%


\global\long\def\Norm{\mathcal{N}}%
\global\long\def\Gam{\textrm{Gam}}%
\global\long\def\InvGam{\textrm{InvGam}}%

\global\long\def\d#1{\ensuremath{\operatorname{d}\!{#1}}}%
\global\long\def\diag{\textrm{diag}}%
\global\long\def\supp{\textrm{supp}}%
\global\long\def\indep{\mathpalette{\independenT}{\perp}}%
\global\long\def\independenT#1#2{\mathrel{\rlap{$#1#2$}\mkern2mu  {#1#2}}}%
\global\long\def\inv{^{\raisebox{.2ex}{${\scriptscriptstyle -1}$}}}%

\global\long\def\R{\mathbb{R}}%

\global\long\def\mba{\bm{a}}%
\global\long\def\mbb{\bm{b}}%
\global\long\def\mbc{\bm{c}}%
\global\long\def\mbd{\bm{d}}%
\global\long\def\mbe{\bm{e}}%
\global\long\def\mbg{\bm{g}}%
\global\long\def\mbh{\bm{h}}%
\global\long\def\mbi{\bm{i}}%
\global\long\def\mbj{\bm{j}}%
\global\long\def\mbk{\bm{k}}%
\global\long\def\mbl{\bm{l}}%
\global\long\def\mbm{\bm{m}}%
\global\long\def\mbn{\bm{n}}%
\global\long\def\mbo{\bm{o}}%
\global\long\def\mbp{\bm{p}}%
\global\long\def\mbq{\bm{q}}%
\global\long\def\mbr{\bm{r}}%
\global\long\def\mbs{\bm{s}}%
\global\long\def\mbt{\bm{t}}%
\global\long\def\mbu{\bm{u}}%
\global\long\def\mbv{\bm{v}}%
\global\long\def\mbw{\bm{w}}%
\global\long\def\mbx{\bm{x}}%
\global\long\def\mby{\bm{y}}%
\global\long\def\mbz{\bm{z}}%
\global\long\def\mbA{\bm{A}}%
\global\long\def\mbB{\bm{B}}%
\global\long\def\mbC{\bm{C}}%
\global\long\def\mbD{\bm{D}}%
\global\long\def\mbE{\bm{E}}%
\global\long\def\mbF{\bm{F}}%
\global\long\def\mbG{\bm{G}}%
\global\long\def\mbH{\bm{H}}%
\global\long\def\mbI{\bm{I}}%
\global\long\def\mbJ{\bm{J}}%
\global\long\def\mbK{\bm{K}}%
\global\long\def\mbL{\bm{L}}%
\global\long\def\mbM{\bm{M}}%
\global\long\def\mbN{\bm{N}}%
\global\long\def\mbO{\bm{O}}%
\global\long\def\mbP{\bm{P}}%
\global\long\def\mbQ{\bm{Q}}%
\global\long\def\mbR{\bm{R}}%
\global\long\def\mbS{\bm{S}}%
\global\long\def\mbT{\bm{T}}%
\global\long\def\mbU{\bm{U}}%
\global\long\def\mbV{\bm{V}}%
\global\long\def\mbW{\bm{W}}%
\global\long\def\mbX{\bm{X}}%
\global\long\def\mbY{\bm{Y}}%
\global\long\def\mbZ{\bm{Z}}%
\global\long\def\mbalpha{\bm{\alpha}}%
\global\long\def\mbbeta{\bm{\beta}}%
\global\long\def\mbdelta{\bm{\delta}}%
\global\long\def\mbepsilon{\bm{\epsilon}}%
\global\long\def\mbchi{\bm{\chi}}%
\global\long\def\mbeta{\bm{\eta}}%
\global\long\def\mbgamma{\bm{\gamma}}%
\global\long\def\mbiota{\bm{\iota}}%
\global\long\def\mbkappa{\bm{\kappa}}%
\global\long\def\mblambda{\bm{\lambda}}%
\global\long\def\mbmu{\bm{\mu}}%
\global\long\def\mbnu{\bm{\nu}}
\global\long\def\mbomega{\bm{\omega}}%
\global\long\def\mbphi{\bm{\phi}}%
\global\long\def\mbpi{\bm{\pi}}%
\global\long\def\mbpsi{\bm{\psi}}%
\global\long\def\mbrho{\bm{\rho}}%
\global\long\def\mbsigma{\bm{\sigma}}%
\global\long\def\mbtau{\bm{\tau}}%
\global\long\def\mbtheta{\bm{\theta}}%
\global\long\def\mbupsilon{\bm{\upsilon}}%
\global\long\def\mbvarepsilon{\bm{\varepsilon}}%
\global\long\def\mbvarphi{\bm{\varphi}}%
\global\long\def\mbvartheta{\bm{\vartheta}}%
\global\long\def\mbvarrho{\bm{\varrho}}%
\global\long\def\mbxi{\bm{\xi}}%
\global\long\def\mbzeta{\bm{\zeta}}%
\global\long\def\mbDelta{\bm{\Delta}}%
\global\long\def\mbGamma{\bm{\Gamma}}%
\global\long\def\mbLambda{\bm{\Lambda}}%
\global\long\def\mbOmega{\bm{\Omega}}%
\global\long\def\mbPhi{\bm{\Phi}}%
\global\long\def\mbPi{\bm{\Pi}}%
\global\long\def\mbPsi{\bm{\Psi}}%
\global\long\def\mbSigma{\bm{\Sigma}}%
\global\long\def\mbTheta{\bm{\Theta}}%
\global\long\def\mbUpsilon{\bm{\Upsilon}}%
\global\long\def\mbXi{\bm{\Xi}}%
\global\long\def\mbzero{\bm{0}}%
\global\long\def\mbone{\bm{1}}%
\global\long\def\mbtwo{\bm{2}}%
\global\long\def\mbthree{\bm{3}}%
\global\long\def\mbfour{\bm{4}}%
\global\long\def\mbfive{\bm{5}}%
\global\long\def\mbsix{\bm{6}}%
\global\long\def\mbseven{\bm{7}}%
\global\long\def\mbeight{\bm{8}}%
\global\long\def\mbnine{\bm{9}}%

\title{Toward a Better Monitoring Statistic for Profile Monitoring via Variational Autoencoders}

  \author{Nurretin Dorukhan Sergin, Hao Yan   \thanks{
    The authors gratefully acknowledge the support from \textit{NSF DMS 1830363. If you have any questions, please feel free to contact haoyan@asu.edu.}} \\
    School of Computing and Augmented Intelligence \\
    Arizona State University\\
}
\maketitle
\begin{abstract}
Variational autoencoders have been recently proposed for the problem of process monitoring. While these works show impressive results over classical methods, the proposed monitoring statistics often ignore the inconsistencies in learned lower-dimensional representations and computational limitations in high-dimensional approximations. In this work, we first manifest these issues and then overcome them with a novel statistic formulation that increases out-of-control detection accuracy without compromising computational efficiency. We demonstrate our results on a simulation study with explicit control over latent variations, and a real-life example of image profiles obtained from a hot steel rolling process.
\medskip{}

\end{abstract}

\section{Introduction}

\label{sec:introduction} Profile monitoring has attracted a growing
interest in the literature in the past decades for its ability to
construct control charts with much better representations for certain
types of process measurements \parencite{Woodall2004-bp,Woodall2007-xs,Maleki2018-uo}.
A profile can be defined as a functional relationship between the
response variables and explanatory variables or spatiotemporal coordinates.
In this work, we focus on the case where the profiles generated from
the process are high-dimensional, \ie, the number
of such explanatory variables or spatiotemporal coordinates are large.
Specifically, we focus on the case where profiles are observed in a high-dimensional space, but profile-to-profile variation lies on a nonlinear low-dimensional manifold.
Our motivating example of such high-dimensional profiles is presented in \ref{fig:Rolling}
below, in which we exhibit a sample of surface defect image profiles
collected from a hot steel rolling process.

\begin{figure}[t]
\includegraphics[width=0.5\textheight]{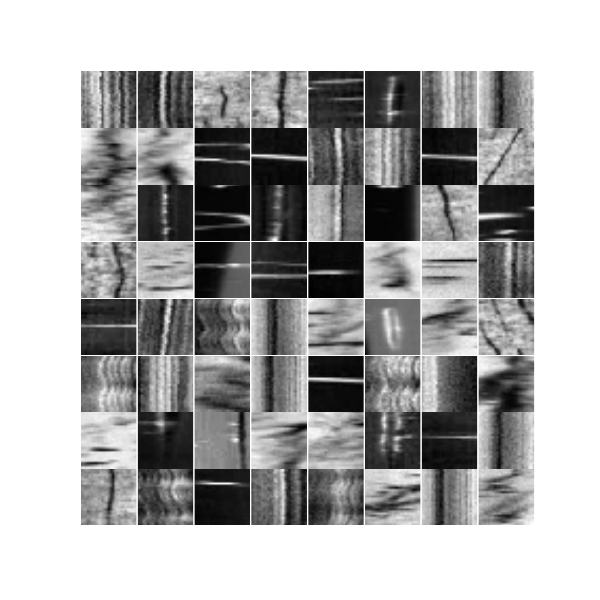} \caption{A collection of 64 by 64 image profiles taken from a hot steel rolling
process.}
\label{fig:Rolling}
\end{figure}

In literature, profile monitoring techniques can be categorized by their assumptions on the type of functional relationship. 
For example, linear profile monitoring techniques assumed that the profiles can be represented by a linear function. 
The idea is to extract the slope and the intercept from each profile and monitor its coefficients \parencite{zhu2009monitoring}.
Regularization techniques can also be used in linear profile estimation. For example, \textcite{zou2012lasso} utilizes a multivariate linear regression model for profiles with the LASSO penalty and use the regression coefficients for Phase-II monitoring. 
However, the linear assumption can be quite limiting. To address this challenge, nonlinear parametric models are normally proposed \parencite{Williams2007-ty,Jensen2009-tu,Noorossana2011-oj,Maleki2018-uo}.
These models assume an explicit family of parameterized functions
and, their parameters are estimated via nonlinear regression. In both
cases, the drawback of both linear and nonlinear parametric models
is that they assume the parametric form is known beforehand, which
might not always be the case in practice.

Another large body of profile monitoring research focuses on the type
of profiles where the basis of the representation is assumed to be
known, but the coefficients are unknown. For instance, to monitor
smooth profiles, various non-parametric methods based on local kernel
regression \parencite{zou2008monitoring,qiu2010nonparametric} and
splines \parencite{chang2010statistical,Yan2018-ux} are developed. To monitor
the non-smooth waveform signals, a wavelet-based mixed effect model
is proposed \parencite{paynabar2011characterization}. However, for
all the aforementioned methods, it is assumed that the nonlinear variation
pattern of the profile is well captured by a known basis or kernel.
Usually, there is no guidance on selecting the right basis of the
representation for the original data and it requires
many trials and errors to find the right basis.

In the case that the basis of HD profiles are not known, dimensionality
reduction techniques are widely used. Principal component analysis
(PCA) is arguably the most popular method in this context for profile monitoring because of its simplicity, scalability, and good data compression capability. In \textcite{liu1995control}, PCA is proposed
to reduce the dimensionality of the streaming data where $T^{2}$ and
$Q$ charts are constructed to monitor the extracted representations
and residuals, respectively. To generalize PCA methods to monitor
the complex correlation among the channels of multi-channel profiles,
\textcite{paynabar2015change} propose a multivariate functional PCA
method and apply change point detection methods on the function coefficients.
Along this line, tensor-based PCA methods are also proposed for multi-channel
profiles, examples including uncorrelated multi-linear PCA \parencite{paynabar2013monitoring}
and multi-linear PCA \parencite{grasso2014profile}, and  
various tensor-based decomposition methods \parencite{yan2015image}.

The main limitation of all the aforementioned PCA-related methods
is that the expressive power of linear transformations is very limited.
Furthermore, each principal component represents a global variation
pattern of the original profiles, which is not efficient at capturing
the local spatial correlation within a single profile. Therefore,
PCA requires much larger latent-space dimensions than the dimension
of the actual latent space, yielding a sub-optimal and overfitting-prone
representation. This phenomenon hinders profile monitoring performance.

A systematic discussion of this issue is articulated in \parencite{Shi2016-tg}.
In that work, the authors identify the problems associated with assuming
a closeness relationship in the subspace that is characterized by
Euclidean metrics. They successfully observe that the intra-sample
variation in complex high-dimensional corpora may lie on a nonlinear
manifold as opposed to a linear manifold, which is assumed by PCA
and related methods. However, the authors only focus on applying manifold
learning for Phase-I analysis, while the Phase-II monitoring procedure
is not touched upon.

In recent years, we observe a surge in deep learning-based solutions to the problem. For instance, deep autoencoders have been proposed for profile monitoring for Phase-I
analysis in \textcite{Howard2018-op}. In another work, \textcite{Yan2016-wa} compared
the performance of contractive autoencoders and denoising autoencoders
for Phase-II monitoring. \textcite{Zhang2018-js} proposed a denoising
autoencoder for process monitoring. Aside from deterministic deep
neural networks, only three works \parencite{wang2019systematic,Zhang2019-lu,lee2019process}
proposed to use deep probabilistic latent variable models, specifically,
variational autoencoders (VAE), for Phase-II monitoring. All the monitoring statistics in those works differ slightly, but
they are all extensions of the classic $T^{2}$ and $Q$-charts of
PCA. We argue that there is room for improvement for the monitoring
statistic formulations in those works for several reasons, especially
when high-dimensional profiles are considered. In this work, we propose
a new monitoring statistic formulation to address this issue.

The contributions of this work are as follows:
\begin{itemize}
\item We compare the existing monitoring statistics proposed by previous works on VAE-based monitoring and unify them into the latent-space and residual-space monitoring statistics. We also prove the mathematical equivalency of these statistics with the classical $T^{2}$ and $Q$-charts of PCA in the linear setting.
\item We highlight an important shortcoming of neural network-based encoders and how it negatively impacts the efficiency of statistics that are derived exclusively from learned latent representations. We demonstrate this on a carefully designed simulation study with explicit control over the actual latent variations.
\item We explain why residual-space monitoring statistics can cover most types of process drifts in both conceptual illustration and real simulation study. 
\item We propose two approximations on the residual-space monitoring statistics leveraging on the first-order and second-order Taylor expansion that strikes a better balance between detection accuracy and computational feasibility than previously proposed similar statistics.
\item We support our claims on both simulation and real-life case study profile datasets.
\end{itemize}
The rest of the paper is organized as follows: \ref{sec:Background}
first introduces variational autoencoders and reviews traditional
$T^{2}$ and $Q$ charts of PCA as well as the existing monitoring
statistics proposed for VAE. \ref{sec:methodology} introduces our proposed monitoring statistic formulation and the rationale behind how it tackles the shortcomings of existing formulations. \ref{sec:Simulation-Study-Analysis} introduces the simulation process used in this work as well as the manifestations of the aforementioned shortcomings.
Finally, \ref{sec:case-study} demonstrates the advantages of the
proposed methodology on a real-life case study,
using images from a hot-steel rolling process.

\section{Background \label{sec:Background}}
In this section, we review the variational autoencoder (VAE) in \ref{sec:bckgrnd:lvms}. We will then review the $T^2$ and $Q$ statistics for PCA methods in \ref{sec:bckgrnd:ReviewPCA}. Finally, we will briefly review the existing works profile monitoring works utilizing the VAE in \ref{sec:bckgrnd:critique}.

\subsection{Variational Autoencoders \label{sec:bckgrnd:lvms}}

We will first review the variational autoencoder (VAE), which was first introduced by \textcite{Kingma2013-dl}. VAE soon became one of the most prominent probabilistic models in the literature. 
The Gaussian factorized latent variable model perspective of VAEs is crucial to understand the role of this model in the context of profile monitoring. 
This is why we begin with an introduction to latent variable modeling.

Let us assume we observe samples $\mbx\in\R^{d}$ in a high-dimensional space, generated by a multivariate random process that can be described by the density function $ p(\mbx) $. We also believe that there is redundancy in this observation and sample-to-sample variation can be explained well by a latent representation $\mbz\in\R^{r}$, where the latent dimension $ r \ll d $. 
Latent variable models are powerful tools to model such complex distributions. The joint density $ p(\mbx,\mbz) $ is factorized into the distribution of the latent variables $\pz$ and the conditional distribution of observed variables given latent variables $p(\mbx\g\mbz)$. A typical example of latent variable models is when the joint distribution is Gaussian factorized as in \ref{eq:gaussian-factorized}. 
\begin{equation}
\begin{split}\pz & =\Norm(\mbz;0,\mbI_{r})\\
\decoding & =\Norm(\mbx;\mu_{\mbtheta}(\mbz),\sigma^{2}\mbI_{d})\\
p_{\mbtheta}(\mbx,\mbz) & =\decoding\pz
\end{split}
\label{eq:gaussian-factorized}
\end{equation}
In the above formulation, the function $\mu_{\mbtheta}\colon\R^{r}\to\R^{d}$
is a function parameterized by $\mbtheta$, which describes
the relationship between the latent variables and the mean of the
conditional distribution. The Gaussian prior $\pz$ is typically chosen
to be standard multivariate Gaussian distribution to avoid degenerate
solutions \parencite{roweis1999unifying} and conditional covariance
is typically assumed to be isotropic $\sigma^{2}I_{d}$ to avoid ill-defined
problems. The aim is to approximate the true density $p_{\mbtheta}(\mbx)\approx p(\mbx)$
and this approximation can be obtained through marginalization: 
\[
p_{\mbtheta}(\mbx)=\int p_{\mbtheta}(\mbx,\mbz)d\mbz
\]

A famous member of the family of models described above is the probabilistic
principal component analysis (PPCA) \parencite{tipping1999probabilistic}.
The parameters are optimized via a maximum likelihood estimation framework
and it can be solved analytically since the function $\mu_{\mbtheta}$ is a simple
linear transformation. This enables reusing analytical results from
solutions to the classical PCA problem. The assumption of PPCA that
the latent and observed variables have a strictly linear relationship
is restrictive. In real-world processes, this relationship is likely
highly nonlinear. Deep latent variable models are a marriage of
deep neural networks and latent variable models that aim to solve
this problem. Deep learning has enjoyed a tremendous resurgence in
the last decade due to their superior performance that was unprecedented
for many tasks such as image classification \parencite{krizhevsky2012imagenet},
machine translation \parencite{bahdanau2014neural}, and speech recognition
\parencite{amodei2016deep}. In theory, under sufficient conditions,
a two-layer multilayer perceptron can approximate any function on
a bounded region \parencite{cybenko1989approximation,Hornik1991-li}.
However, growing the width of shallow networks exponentially
for arbitrarily complex tasks is not practical. It has been shown
that deeper representations can often achieve better expressive power
than shallow networks with fewer parameters due to the efficient reuse
of the previous layers \parencite{eldan2016power}.

VAE is arguably the most foundational member of the deep latent variable
model family. The main difference between PPCA and VAE is that VAE
replaces the linear transformation with a high-capacity deep neural
network (called \textit{generative} or \textit{decoder}). This is
powerful in the sense that, along with a general-purpose prior $\pz$,
deep neural networks can transform such prior to model a wide variety
of densities to model the training data \parencite{kingma2019introduction}.
Unlike PPCA, these models will not have analytical solutions due to
the complex nature of the neural network used. Like most other deep
learning models, their parameters are often optimized via variants of stochastic gradient 
descent optimizers. The problem becomes even harder given
that the posterior $\decoding$ takes meaningful values
only for a small sub-region within the latent space $\R^{r}$. This makes sampling
from the prior $\pz$ to estimate the likelihood prohibitively expensive.
Both models work around this problem using the importance sampling
framework \parencite[532]{bishop2006pattern}, where they introduce
another network (called \textit{recognition} or \textit{encoder})
to approximate a proposal distribution $\encoding$ ---parametrized
by $\mbphi$--- which aims to sample latent variables from
a much smaller region that is more likely to produce higher posterior
densities for a given input $\mbx$. The encoder is modeled as another Gaussian distribution $\encoding = \Norm(\mbz;\mu_{\mbphi}(\mbx),\sigma_{\mbphi}(\mbx))$ where the mean and standard deviation of the proposal distribution are inferred via high capacity neural networks $\mu_{\mbphi}$ and $\sigma_{\mbphi}$, respectively.

One important output of a trained VAE is the likelihood estimator.
Once the two networks are trained, the log-likelihood $\log\ptheta(\mbx)$
can be approximated by a Monte Carlo sampling procedure with $L$
iterations \parencite[30]{kingma2019introduction}: 
\begin{equation}
\log\ptheta(\mbx)\approx\log\frac{1}{L}\sum_{l=1}^{L}\frac{\ptheta(\mbx,\mbz^{(l)})}{\qphizgivenx{\mbz^{(l)}}{\mbx}}.\label{eqn:SummationLL}
\end{equation}

However, the Monte Carlo sampling procedure is shown to be computationally inefficient and the evidence lower bound (ELBO),
which is deemed a proxy to the likelihood, is often used as the objective to be optimized. 
\begin{equation}
\begin{split}\text{ELBO} & \triangleq\log\left(p(\mbx)\right)-\KL{\encoding}{q^{*}(\mbz|\mbx)}\\
 & =\E_{\mbz\sim q_{\mbtheta}}\log\decoding+\KL{\encoding}{p(\mbz)},
\end{split}
\label{eqn:VAELoss}
\end{equation}
In the equation above, $\KL{\cdot}{\cdot}$ denotes the Kullback-Leibler divergence
(KLD) between two distributions. The left-hand side is the quantity
of interest, while the right-hand side is the tractable expression
that guides the updating of parameters $\mbtheta,\mbphi$ in an end-to-end
fashion.

\subsection{Review of $\protect\Tsq$ and $Q$ Statistics in PCA \label{sec:bckgrnd:ReviewPCA}}
We will then review the profile monitoring statistics in the principal component analysis (PCA). 
Profile monitoring via PCA is typically done using the $\Tsq$ and $Q$ statistics \parencite{Chen2004-px}.
The $Q$ statistic for PCA is defined as the reconstruction error
between the observed profile $\mbx$ and the reconstructed profile $\tilde{\mbx}$.
The geometric interpretation of $ Q $ statistics is that it quantifies how far the sample is away from the
learned subspace of in-control samples. $\Tsq$ statistics on the other hand, quantifies the shift along the directions of the most dominant principal components.

The $\Tsq$ statistic and $Q$ statistic for PCA are defined formally as follows:
\begin{equation}
\begin{split}Q(\mbx) & =\gg\mbx-\tilde{\mbx}\gg^{2}\\
\Tsq(\mbx) & =\mbz^{\top}\mbSigma\inv_{r}\mbz=\mbx^{\top}\mbW_{r}\mbSigma\inv_{r}\mbW_{r}^{\top}\mbx,
\end{split}
\label{eqn: QTPCA}
\end{equation}
where matrix $\mbW_{r}$ is the loading matrix, and $\mbSigma\inv_{r}$
is the inverse of the covariance matrix when only the first $r$ principal
components are kept. There are various methods to choose $r$ such
as fixing the percentage of variation explained \parencite[41]{Chiang2001-nu}.

For processes with relatively small latent and residual dimensionality,
the upper control limits of these statistics for the $\alpha$\% Type-1
error tolerance is constructed by employing the normality assumptions
of PPCA \parencite[43-44]{Chiang2001-nu}. However, using such measures
for high-dimensional nonlinear profiles is prohibitively error-prone
as both $r$ and $d$ will be much higher than the assumptions of
chi-square distribution can tolerate. As an alternative, non-parametric
methods are typically used to estimate these limits, such as simple percentiles or kernel density estimators.

\subsection{Review of Previously Proposed Monitoring Statistics
Proposed for VAE \label{sec:bckgrnd:critique} }

In this section, we will briefly review several proposed monitoring statistics for variational autoencoders (VAE).s Three works have recently considered
VAE for process monitoring, all of which propose different statistic
formulations for monitoring. \textcite{Zhang2019-lu} propose $H^{2}$, which is basically the
Mahalanobis distance of the mean of the proposal distribution from standard Gaussian distribution. 
\begin{equation}
H^{2}=\mu_{\mbphi}(\mbx)^{\top}\mu_{\mbphi}(\mbx)
\end{equation}

In another work, \textcite{lee2019process} propose two statistics: $T^{2}$ and $SPE$. For a given input $\mbx$, a single sample is drawn from the proposal distribution $\mbz^{(l)}\sim\encoding$ which is used reconstruct the input using the generative model $\mbx^{(l)}\sim p_{\mbtheta}(\mbx\g\mbz^{(l)})$. The proposed test statistics in this work can be formalized as follows:
\begin{equation}
\begin{aligned}T^{2} & =(\mbz^{(l)}-\bar{\mbz})^{\top}S_{\mbz}\inv(\mbz^{(l)}-\bar{\mbz})\\
SPE & =\gg\mbx^{(l)}-\mbx\gg_{2}^{2},
\end{aligned}
\end{equation}
where $\bar{\mbz}$ and $S_{\mbz}\inv$ are estimated over a single pass of the entire set of in-control samples. In their methodology, these two statistics work in combination and at least one positive decision from either of the two statistics is enough to claim that the process is out-of-control. 

Finally, \textcite{wang2019systematic} propose the $R$ and $D$ statistics by focusing on the two major components of the tractable part of the objective function of VAE shown as in \ref{eqn:VAELoss}. The $D$ statistic is simply the KL divergence between the prior and proposal. For $R$ statistic, like \textcite{lee2019process}, they employ summary statistics over samples from proposal but also claim that sampling size can be fixed to one: 
\begin{equation}
\begin{aligned}D & =\KL{\encoding}{p(\mbz)}\\
R & =\frac{1}{L}\sum_{l=1}^{L}-\log q_{\mbtheta}(\mbx\g\mbz^{(l)}),
\end{aligned}
\label{eq: DR}
\end{equation}

$SPE$ in and $R$ are essentially the same quantities up to a constant, which makes them identical in the context of monitoring. This is why we will refer to them as $SPE/R$ throughout the rest of the paper.

\section{Methodology \label{sec:methodology}} 

In this section, we start by explaining how previously proposed statistics for VAE-based monitoring are modeled as extensions of their PCA-based monitoring counterparts, in \ref{sec:residual-latent-vae-pca}. Then, we will reveal the pitfalls of this extension concerning the behaviors of neural networks in \ref{sec:proposed-statistic}. Against the backdrop of these pitfalls, we will propose a novel monitoring statistic formulation. Lastly, we will outline the implementation details of profile monitoring procedures and neural network architectures we use in this study in \ref{sec:methodology:procedure} and \ref{subsec:Model-Architectures}, respectively. 

\subsection{Relationship of the Monitoring Statistics for VAE and PCA} 
\label{sec:residual-latent-vae-pca}

A common approach in the literature to tackle process monitoring with VAE is to extend the definitions of $ \Tsq $ and $ Q $ statistics of the PCA-based monitoring to VAE. This is done by breaking the tractable portion of \ref{eqn:VAELoss} into two term as follows:

\begin{equation}
Q_{VAE}  = \E_{\mbz\sim q_{\mbtheta}}\log\decoding, 
T^2_{VAE} = \KL{\encoding}{p(\mbz)}.  \label{eq: TQVAE}
\end{equation}

Either these formulations or some variant of them are typically used as the monitoring statistics. To understand the rationale behind this, we will revisit the assumptions of the model described in \ref{eq:gaussian-factorized}.
Let us formally represent an out-of-control distribution as a shift in $p(\mbx)$. 
Since $p(\mbx)=\int p(\mbx\g\mbz)p(\mbz)d\mbz$, we can anticipate two sources: a shift in the latent distribution $\pz$ or a shift in the residual distribution $p(\mbx\g\mbz)$. The two statistics are assumed to be connected to these two sources: 1) a shift in the conditional distribution $ p(\mbx\g\mbz) $ can be detected monitoring $Q_{VAE}=\E_{\mbz\sim q_{\mbtheta}}\log\decoding$ and
2) a shift in the latent distribution $p(\mbz)$, can be detected monitoring $T^2_{VAE}=\KL{\encoding}{p(\mbz)}$. 

This idea is similar to utilizing both $T^2$ and $Q$-charts in the PCA-based method, where both terms play an important role in process monitoring \parencite{kim2003process}. To make this similarity more obvious, we prove that if the same ELBO framework for VAE used above is used for PPCA (see \ref{sec:bckgrnd:ReviewPCA}), we prove the equivalency of $T^{2}$ and $Q$ statistics of PPCA and $T^2_{VAE}$ and $Q_{VAE}$ of VAE in the linear settings.

\begin{prop}
	\label{prop: T2Q} If we use linear decoders for VAE, the models will become the Probabilistic PCA \parencite{tipping1999probabilistic}
	that the prior and decoding functions are normally distributed as: 
	\[
	\begin{split}p(\mbz) & =\Norm(0,\mbI),\\
		\decoding & =\Norm(\mbW\mbz,\sigma^{2}\mbI).\label{eq:Gaussian}
	\end{split}
	\]
	In this case, the encoder can be solved analytically as
	another normal distribution as $\encoding=\Norm(\mu_{\mbphi}(\mbx),\Sigma_{z})$,
	where $\mu_{\mbphi}(\mbx)=\mbM^{-1}\mbW^{\top}\mbx$, $\Sigma_{z}=\sigma^{2}\mbM^{-1}$,
	and $\mbM=\mbW^{\top}\mbW+\sigma^{2}\mbI$. Then, the two monitoring statistics defined in \ref{eq: TQVAE} can be derived as follows:
	\begin{equation}
		\KL{\encoding}{p(\mbz)}=\frac{1}{2}\gg\mu_{\mbphi}(\mbx)\gg^{2}+C_{1},\label{eqn:KL_PPCA}
	\end{equation}
	\begin{equation}
		\E_{\mbz\sim q_{\mbphi}}\log\decoding\propto\gg\mbx-\mbW\mu_{\mbphi}(\mbx)\gg^{2}+C_{2},\label{eqn:E_PPCA}
	\end{equation}
	where $C_{1}$ and $C_{2}$ are constants that do not depend on $x$. The proof is given in \ref{sec:PoofOfPropTQ}.
\end{prop}

Note that the constants do not affect the profile monitoring decision as the control limits will be translated accordingly. Thus, the test statistic $T^2_{VAE}$ and $Q_{VAE}$ for linear decoders (i.e., PPCA) is equivalent to the $T^{2}$-statistic and $Q$-statistic of PCA, respectively. residual-space

Observe that previously proposed formulations mentioned in \ref{sec:bckgrnd:critique} draw inspiration---directly or indirectly---from this framework. Statistics $R$ and $SPE$ are based on the $Q$-statistic. Let us call these \emph{residual-space statistics}, as they rely on the sum of squared differences between the signal itself and its predicted value, \ie, residuals. The statistics $H^{2},T^{2}$ and $D$ are based on the $T^{2}$ of PCA. We call these \emph{latent-space statistics}, as they rely exclusively on latent representations. 

\begin{figure}[t]
	\begin{centering}
		\includegraphics[width=0.9\textwidth]{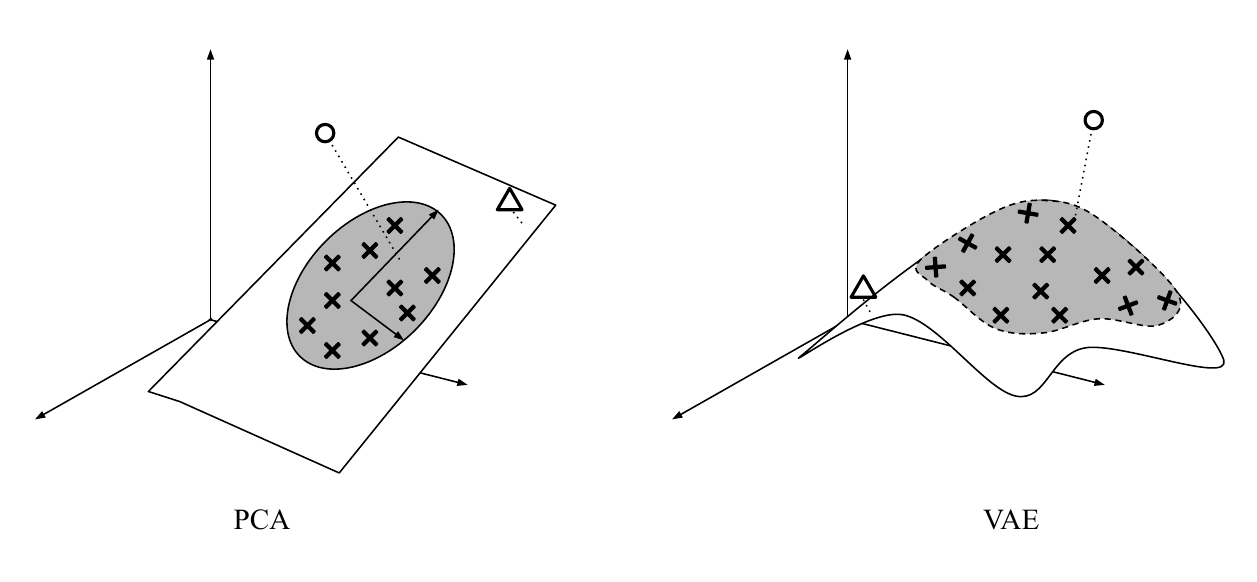}
		\caption{Illustration of the analogy between PCA and VAE. Closed regions describe the lower-dimensional manifold the in-control distribution lies in. The crosses represent the in-control samples observed in Phase-I and the gray region represents the subset of the lower-dimensional manifold where in-control samples are typically sampled from. The observation represented with a circle is typically detected with $ Q $-statistic and the observation represented with a triangle is typically detected with $ \Tsq $-statistic. \label{fig:pcaVSvae}}	
	\end{centering}
\end{figure}

\ref{fig:pcaVSvae} shows a graphical illustration of this analogy of residual-space statistics and latent-space statistics for PCA and VAE. Residual-space statistics quantify the distance of the observed data with respect to the learned linear or nonlinear manifold. The latent-space statistics monitors the distance within the learned manifold. In the linear case (i.e., PCA), this is the Euclidean distance. However, in the nonlinear case (i.e., VAE), this distance should be defined on the nonlinear manifold.

\subsection{Proposed Monitoring Statistic}

\label{sec:proposed-statistic}

\begin{figure}[t]
	\begin{centering}
		\includegraphics[width=0.9\textwidth]{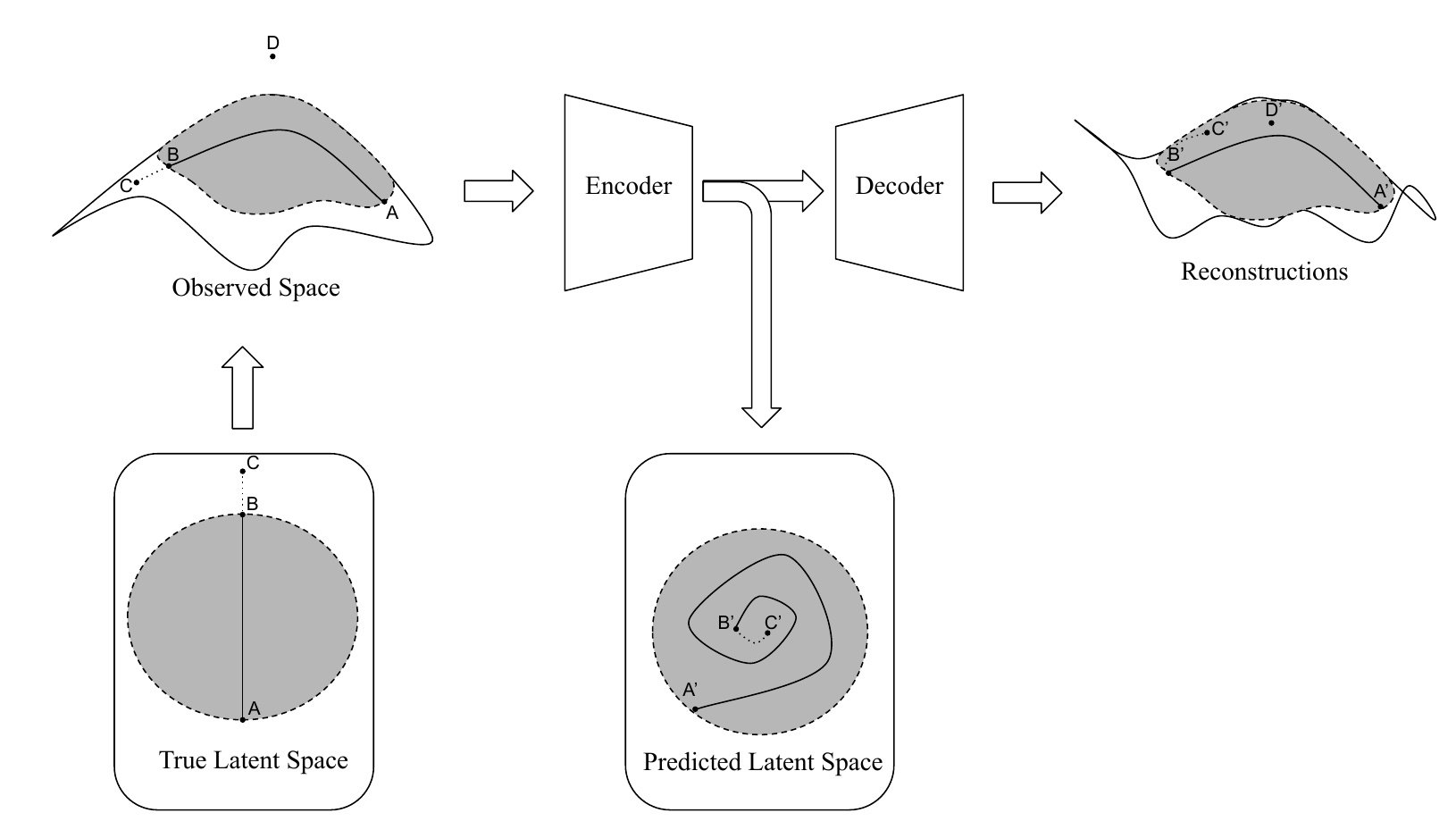}
		\par\end{centering}
	\caption{Illustration of incorrect latent representation phenomena and how process control fails in latent space. \textbf{Bottom left:} The true latent variations of in-control samples are generated from the gray region, which is the probable region. Point A and Point B are extreme values along a dimension of variation. Point C is generated by an out-of-control process with a shift in latent distribution. Point D is generated by an out-of-control process with a shift in the residual distribution. The predicted counterpart of each point is denoted by an apostrophe (\eg, A' for A). \textbf{Top Left:} Observations of true latent variation in the high-dimensional space that lie close to a low-dimensional manifold. \textbf{Top Middle:} The encoder and decoder of VAE trained exclusively with in-control samples (\ie, the gray region in the observed space). \textbf{Bottom Middle:} Incorrectly mapped variation in the predicted latent space where the gray region is the probable region. \textbf{Top Right:} Reconstructions of the variation in high-dimensions, with a failure in extrapolation beyond the in-control region. \label{fig:entang-extrap}}
\end{figure}

In this section, we will first reveal the shortcomings of the previously proposed VAE-based monitoring methodologies we present in \ref{sec:bckgrnd:critique}. This will lead us to the rationale behind the design of our proposed statistic, which is also included in this section after the explanation of the shortcomings.

There are two major pitfalls of the previously proposed methodologies:
\begin{enumerate}
	\item Latent-space statistics $H^{2},T^{2}$ and $D$ or any other formulation that relies exclusively on the latent representation $ \mbz \sim \qphizgivenx{\mbz}{\mbx}$ will be unreliable for process monitoring. Thus, they should be discarded altogether from the monitoring framework since they will likely increase false alarms without contributing to the detection power in any meaningful way.
	\item Residual-space statistic $ SPE $ and $ R $ rely on Monte Carlo sampling. These are not computationally feasible given how expensive calculations are on deep neural networks. An alternative approach is required to stay computationally feasible without sacrificing too much from the estimation quality of these statistics.
\end{enumerate}
We will address these two shortcomings in \ref{sec:unreliable} and \ref{sec:efficiency}. 

\subsubsection{Unreliability of Latent-space Statistics for Deep Autoencoders} \label{sec:unreliable}

First, we focus on the unreliability of latent-space statistics. Let us first start with the case when the shift occurs in the latent distribution (i.e., $\pz$). According to the PCA-VAE analogy illustrated in \ref{fig:pcaVSvae}, latent-space statistics are supposed to catch such shifts, which are represented with triangular points in the same figure. While this may work for PCA-based monitoring, we claim that such an analogy cannot be straightforwardly made for VAE. Here, we will explain the two major reasons why the latent-space statistics failed to capture the change in the latent space: "incorrect" latent representation by the encoder and failure to extrapolate by the decoder. 

The first reason that latent-space statistics should not be used is that neural network-based encoders in autoencoder architectures typically learn "incorrect" latent representation. We illustrate this phenomenon in \ref{fig:entang-extrap}. The line segment ABC illustrates a traversal along a latent dimension. All the samples generated along the line segment AB are sampled from the typical region of the in-control process and their latent representations are contained within the typical region of the predicted space. However, Point C is generated by an out-of-control process where there is a shift in the latent distribution but its mapping incorrectly falls within the probable region. This leads to false evidence which suggests that Point C is unlikely to be generated by an out-of-control process while in reality, it was. 

The reasons as to why incorrect latent representation are learned by deep autoencoders have been studied well in the deep learning literature. Interested readers are encouraged to refer to \textcite{AchilleS18} for a discussion of the properties of ideal latent representations and to \textcite{locatello2018challenging} for a discussion of the challenges around attaining one of these properties, namely, disentanglement. The key takeaway is that it is very likely that we end up with an imperfect mapping, especially with real-life datasets. Consequently, in Phase-II, samples generated by out-of-control processes that are characterized by a shift in the latent distribution will not be mapped consistently to the regions in the latent space, which we consider to be unusual. This will result in an increased type-II error.

A natural question to ask at this point is how we should expect to detect shifts in latent distribution if we cannot rely on latent representations. We argue that the residual-space statistics (i.e., an analog of a $ Q $-chart) would catch such shifts too, even though its original purpose is to catch shifts in the residual space. Our argument is based on another "shortcoming" of neural networks, namely, failure to extrapolate. Deep neural networks approximate well only at a bounded domain defined by where the training set is densely samples from. In the context of our problem, this refers to the high-density region of $ p(\mbx) $, which generated the set of in-control profiles we use in Phase-I. The behavior of the function is unpredictable and often erroneous outside the training domain. In other words, it does not extrapolate well beyond the domain of training samples, which are likely to be coming from out-of-control processes. We refer interested readers to \ref{app:rosenbrock}, where we replicate this phenomenon on a toy example.

This leads to the second reason why the residual-space statistics should be used only: failure to extrapolate by the decoder. A decoder that fails to extrapolate is counter-intuitively helpful for the residual-space statistics since it will struggle with generating profiles that are in the low-density region of the in-control data distribution $ p(\mbx) $. This means that the discrepancy between the true profile and its generated counterpart will be larger for out-of-control profiles than it is for in-control profiles, regardless of the source of the shift. 
Overall, we conclude that the residual-space monitoring statistic would be efficient at detecting changes in the residual space and latent space. We refer the readers to \ref{fig:entang-extrap} for an illustration. 
Point C is generated from a shift in the latent space distribution. However, due to the "incorrect" mapping of the latent distribution, Point C' will still lie in the in-control region of the latent space. There is a significant discrepancy between Point C and reconstruction C', which can be detected by the residual-space statistics. Point D is generated from a shift in the residual space and can be captured by the residual space statistics. In conclusion, the residual-space statistic should be able to catch  changes in both the residual space and latent space.  is that

\subsubsection{Improving the Computational Efficiency of the residual-space Statistics} \label{sec:efficiency}

Now that we established our rationale behind the first shortcoming we claim to reveal, we move onto the second and focus on the previously proposed residual-based statistics: $ SPE $ and $ R $. Both $SPE$ and $ R $ rely on samples from the proposal distribution for the estimation of the expectation. This approach requires a large number of samples to be generated, and thus a large number of the forward passes through the decoder network, which is prohibitively expensive in terms of computation when deployed in real-life systems. To overcome this problem, we propose a Taylor expansion based approximation. First, observe that $\log\decoding\propto\gg x-\mbmu_{\mbtheta}(z)\gg_{2}^{2}+C$ for all $\mbx$ and $\mbz$ because of the common isotropic covariance assumption. The constant $C$ can be discarded as noneffective in terms of control charting because it would only translate the limits and the statistics by the same amount for any given $\mbx$ and $\mbz$. We call the expression $\E_{\mbz\sim q_{\phi}}\gg x-\mbmu_{\mbtheta}(z)\gg_{2}^{2}$ as the expected reconstruction error (ERE). The Taylor expansion for the first-order and second-order moment of ERE given the random variable $\mbz\sim\encoding$ can be derived analytically.
\begin{prop}
\label{prop:taylor-exp}
Assume that a VAE is trained with in-control samples. The training results in the mean and diagonal covariance estimators of the proposal distribution as well as the mean estimator of the condition distribution which are denoted by $ \mbmu_{\mbphi} $,$ \mbsigma_{\mbphi} $, and $ \mbmu_{\mbtheta} $, respectively. The first and second-order Taylor Expansion (denoted by $ERE_{1}$
and $ERE_{2}$ respectively) for the function $\E_{\mbz\sim q_{\phi}}\gg \mbx-\mbmu_{\mbtheta}(z)\gg_{2}^{2}$
given the random variable $\mbz\sim\encoding=\Norm(\mu_{\mbphi}(\mbx),\mbsigma_{\mbphi}(\mbz))$ and where the conditional $\decoding=\Norm(\mu_{\mbphi}(\mbx),\diag(\mbsigma_{\mbphi}(x)))$
can be derived analytically as:
\begin{equation}
ERE_{1}=\gg\mbx-\mbmu_{\mbtheta}(\mbmu_{\mbphi}(\mbx))\gg_{2}^{2}\label{eq:ere-1}
\end{equation}
\begin{equation}
ERE_{2}=\gg\mbx-\mbmu_{\mbtheta}(\mbmu_{\mbphi}(\mbx))\gg_{2}^{2}+\frac{1}{2}\mathrm{tr}(\mathbf{H}_{z}\diag(\mbsigma_{\mbphi}(x)))\label{eq:ere-2}
\end{equation}
where $\mathbf{H}_{z}$ is the Hessian of the function $\gg x-\mbmu_{\mbtheta}(z)\gg_{2}^{2}$
with respect to $\mbz$. The derivation is provided in \ref{app:ere}.
\end{prop}

Given a trained VAE, $ERE_{1}$ can be computed efficiently by a single forward pass of the new profile from the pass $\mbx$ through $\mbmu_{\mbphi}$ and $\mbmu_{\mbtheta}$ successively and calculating the squared prediction error, without the need for any sampling. $ERE_{2}$ requires the additional computation of the diagonal of the Hessian $\mathbf{H}_{z}$ and a relatively less expensive trace operation since the covariance is diagonal. Both $ERE_{1}$ and $ERE_{2}$ are residual-based statistics that are accurate and efficient to compute, which addresses the two shortcomings we mentioned at the beginning of this section. In our experiments, we will evaluate the effectiveness of both of these statistics in comparison to previously proposed monitoring statistics for VAE.

\subsection{Profile Monitoring Procedure}

\label{sec:methodology:procedure} A typical profile monitoring follows
two phases: Phase-I analysis and Phase-II analysis. Phase-I analysis
focuses on understanding the process variability by training an appropriate
in-control mode and selecting an appropriate control limit. In our
case, Phase-I analysis results in a trained model (i.e., an encoder
and a decoder) and an Upper Control Limit (UCL) to help set up the
control chart for each of the monitoring statistics. In Phase-II,
the system is exposed to new profiles generated by the process in
real-time to decide whether these profiles are in-control or out-of-control.
Our experimentation plan, outlined below, is formulated to emulate
this scenario to effectively assess the performance of any combination
of a model, a test statistic, and a disturbance scenario to generate
the out-of-control samples.
\begin{itemize}
\item Obtain in-control dataset $\dataset$ and partition it into train,
validation and test sets $\dataset^{trn}$, $\dataset^{val}$, $\dataset^{tst}$.
\item Train VAE using samples from $\dataset^{trn}$.
\item Calculate test statistic for all $\mbx\in\dataset^{val}$ and take
it's \nth{95} percentile as the UCL.
\item Start admitting profiles online from the process. Calculate test statistic
using the trained VAE. If the test statistic is over UCL, identify the
sample as out-of-control.
\end{itemize}
We train 10 different model instances with different seeds to account
for inherent randomness due to the weight initialization of deep neural
networks.

\subsection{Neural Network Architectures and Training \label{subsec:Model-Architectures}}

In this work, we use convolutional neural networks for the encoders
and decoders in our VAE model to represent the spatial neighborhood
structures of the profiles. Introduced in \textcite{lecun1989backpropagation},
convolutional layers have enabled tremendous performance increase
in certain neural network applications where the data is of a certain
spatial neighborhood structure such as images or audio waveform. They
exploit an important observation of such data, where the learner should
be equivariant to translations. This is an important injection of
inductive bias into the network that largely reduces the number of
parameters compared to the fully connected network by the use of parameter
sharing. It eventually increases the statistical learning efficiency,
especially for small samples. It must be noted, however, convolutional
layers are not equivariant to scale and rotation as they are to translation.
Knowing what sort of inductive biases is injected into these layers
is important for the understanding of disentanglement, which we will
introduce later in this paper.

We use the encoder-decoder structure outlined in \ref{tab:model-architectures}.
The layers used that builds the model architectures used in this study
are summarized as follows:
\begin{itemize}
\item C($O,K,S,P$): Convolutional layer with arguments referring to the
number of output channels $O$, kernel size $K$, stride $S$ and
size of zero-padding $P$.
\item CT($O,K,S,P$): Convolutional transpose layer with arguments referring
to the number of output channels $O$, kernel size $K$, stride $S$,
and size of zero-padding $P$.
\item FC($I,O$): Fully connected layer with arguments referring to input
dimension $I$ and output dimension $O$.
\item A(): Activation function. Leaky ReLU with a negative slope of $0.2$.
\end{itemize}
Here, C(), CT(), and FC() are considered the linear transformation
layers while R(), LR(), and S() are considered the nonlinear activation
layers. Strided convolutions can be used to decrease the spatial dimensions
in the encoders. Pooling layers are typically not recommended in autoencoder-like
architectures \parencite{radford2015unsupervised}. Convolutional
transpose layers are used to upscale latent codes back to ambient
dimensions.

The sequential order of the computational graphs used for this study
is summarized in \ref{tab:model-architectures}. The architecture choice is directly based on the encoder-decoder architecture that was used in \textcite{higgins2017beta}, except that we use Leaky ReLU with a negative slope of 0.2 as the activation, which is advised in \textcite{radford2015unsupervised} for better gradient flow. The encoder  outputs $2r$ nodes, which is a concatenation of the inferred posterior mean $\mbmu_{\mbphi}(\mbx)$ and variance $\diag(\mbsigma(\mbx))$, both are of length $r$. 
The number of epochs per training is fixed at $1000$, and the learning rate and batch size are fixed at $0.001$ and $64$, respectively, both are chosen empirically to guarantee a meaningful convergence. Adam algorithm is used for first-order gradient optimization with parameters $(\beta_{1,}\beta_{2})=(0.9,0.999)$ as advised in \textcite{KingmaB14}. The model checkpoint is saved at every epoch where a better validation loss is observed. The latest checkpoint is used as the final model.

In our experiments, the architecture and the training conditions described above are optimized with respect to the convergence performance of the VAE objective on the in-control dataset. This is because in real life, the practitioner will not have access to out-of-control samples. Consequently, the same setting worked well for both the simulation dataset and the case study dataset we consider in this paper. This gives us confidence that the selection is robust from one set to the other. However, a different dataset might benefit from adjustments to the above conditions. The adjustments should be based on monitoring the convergence of the VAE objective, as the procedure will benefit from a better approximated in-control distribution.

We would like to emphasize that even we focus only on the image profiles in our paper by the convolutional architectures, which will be introduced to the readers in the upcoming  simulation and case study sections, the monitoring statistics we propose in \ref{eq:ere-1} and \ref{eq:ere-2} can be applied to other profiles as well, which will be left as the future work.

\begin{table}[!t]
\global\long\def\arraystretch{1.3}%
 \caption{Architecture details of deep neural networks used in this study\label{tab:model-architectures}}

\centering{}%
\begin{tabular}{l>{\raggedright}p{0.8\textwidth}}
\toprule 
Module & Architecture\tabularnewline
\midrule 
Encoder & C(32, 4, 2, 1) - A() - C(32, 4, 2, 1) - A() - C(64, 4, 2, 1) - A() - C(64,
4, 2, 1) - A() - C(64, 4, 1, 0) - FC(256, $2r$)\tabularnewline
Decoder & FC($r$, 256) - A() - CT(64, 4, 0, 0) - A() - CT(64, 4, 2, 1) - A() - C(32,
4, 2, 1) - CT(32, 4, 2, 1) - A() - CT(1, 4, 2, 1)\tabularnewline
\bottomrule
\end{tabular}
\end{table}

\section{Simulation Study Analysis and Results \label{sec:Simulation-Study-Analysis}}

In this section, we will evaluate the proposed methodology via a simulation study. We will first test our claims we make in \ref{sec:proposed-statistic} in a controlled environment over the data generating process as  described in \ref{sec:simsetting}. For every experiment mentioned in this section, we follow the procedure outlined in \ref{sec:methodology:procedure} and we use VAE models
with the architecture described in \ref{subsec:Model-Architectures}.

We will then illustrate the incorrect mapping of the latent space and the extrapolation issue in \ref{sec:simstudy:recognition} and \ref{sec:simstudy:generator} under this controlled experiment.

\subsection{Simulation Setup}
\label{sec:simsetting} 

We first evaluate the performance of the deep latent variable models in a simulation setting where we have explicit control over the latent variations. 
The simulation procedure produces 2D structured point clouds that resemble the scanned topology of a dome.

Let each pixel on a $64$ by $64$ grid be denoted by a tuple $\mbp=(p_{0},p_{1})$.
The values of the tuples stretch from $0$ to $1$, equally spaced, left to right and bottom-up. Each tuple takes a value based on its location through a function $\mbp\mapsto f(\mbp;c,r)+\epsilon$, where $\epsilon\sim\Norm(0,1\times10^{-2})$ is i.i.d Gaussian noise.
The function $f$ is parameterized by the horizontal location of the dome $c$, and the radius of the base of the dome $r$. The vertical location of the dome on the 2D surface is fixed at the vertical center of the surface. Given any parameter set $\{c,r\}$, each pixel $\mbp$ can be evaluated with the following logic: 
\begin{equation}
\begin{split}g(\mbp;c,r) & =1-\frac{(p_{0}-c)}{r}^{2}-\frac{(p_{1}-0.5)}{r}^{2}\\
f(\mbp;c,r) & =\begin{cases}
\sqrt{g(\mbp;c,r)} & \mbox{if }g(\mbp;c,r)\geq0\\
0 & \mbox{if }g(\mbp;c,r)<0
\end{cases}
\end{split}
\label{eq:gasketfun}
\end{equation}

The samples are best visualized as grayscale images, as shown in \ref{fig:gasketgrid} below. 

\begin{figure}[t]
\centering{} \includegraphics[width=0.5\linewidth]{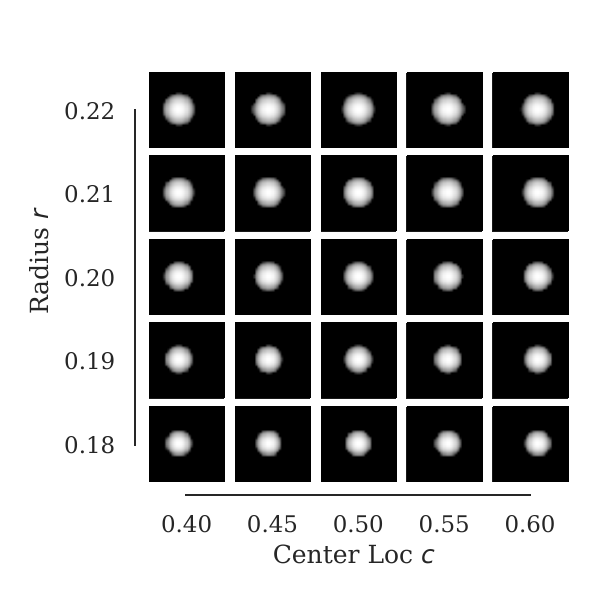}
\caption{Dome profiles depicted as grayscale images simulated with radius and center location they coincide with on the axes.\label{fig:gasketgrid}}
\end{figure}

The processes that generate the latent variations of in-control domes are defined as Gaussian distributions:
\begin{equation}
\begin{split}c\sim\Norm(0.5,1\times10^{-2})\\
r\sim\Norm(0.2,6.25\times10^{-4})
\end{split}
\end{equation}

As our out-of-control scenarios consider the following four distribution shifts in which $\delta$ denotes the intensity of the shift:
\begin{itemize}
\item \textbf{Location shift:} the mean of the process that generates $c$
is altered by an amount $\delta$ as in 
\[
c\sim\Norm(0.5+\delta\times10^{-2},1\times10^{-2})
\]
\item \textbf{Radius shift:} the mean of the process that generates $a$
is perturbed by an amount $\delta$ as in
\[
r\sim\Norm(0.2+\delta\times10^{-4},6.25\times10^{-4})
\]
\item \textbf{Mean shift}: all the pixels are added an additive disturbance
$\delta$ as in 
\[
f(\mbp;c,r)\leftarrow f(\mbp;c,r)+\delta
\]
\item \textbf{Magnitude shift:} all the pixels are added a multiplicative
disturbance $\delta$ as in 
\[
f(\mbp;c,r) \leftarrow f(\mbp;c,r)*\delta
\]
\end{itemize}

Note that the location shift and radius shift represent disturbances in latent distribution $p_{\delta}(\mbz)$. 
The other two cases, mean shift and magnitude shift, represent disturbances in the conditional distribution $p_{\delta}(\mbx\g\mbz)$.

We generate the training, validation, and testing sets for in-control domes as well as a set of each out-of-control scenario above.
All sets have exactly 500 distinct samples.
We generate these sets once, fix them, and use them for the analyses in the subsequent sections.

\subsection{On the Incorrect Mapping of Latent Representations by the Encoder
\label{sec:simstudy:recognition}}
\begin{figure}[t]
	\subfloat[Fixed $r$, varying $c$\label{fig:Fixed-r-varying-c}]{\centering{}\includegraphics[scale=0.5]{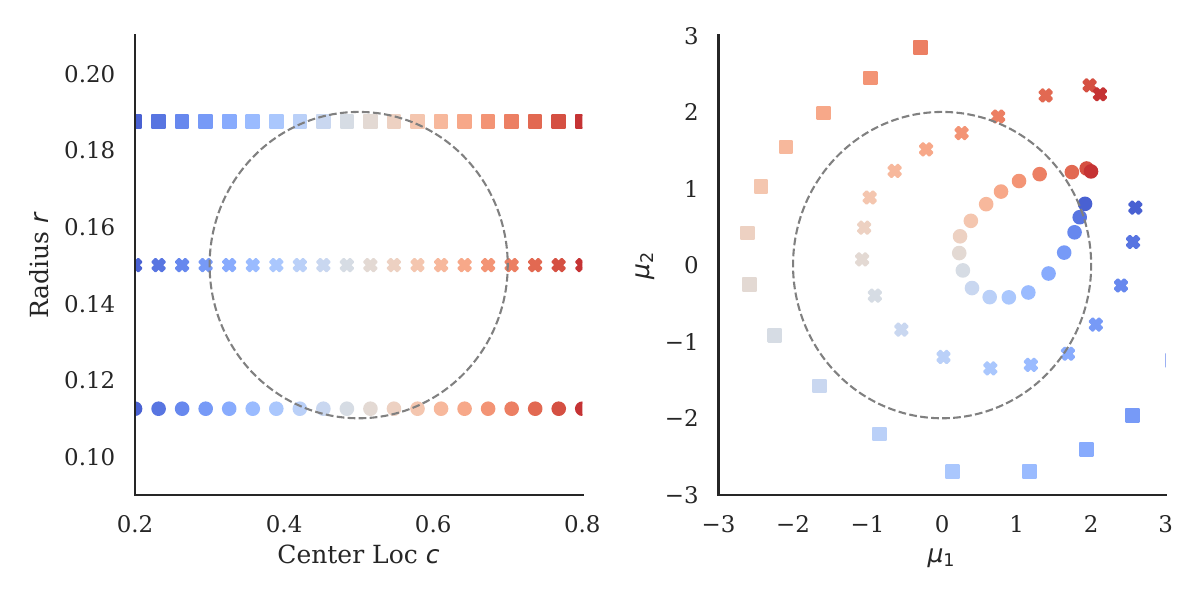}}
	
	\subfloat[Fixed $c$, varying $r$\label{fig:Fixed-c-varying-r}]{\begin{centering}
			\includegraphics[scale=0.5]{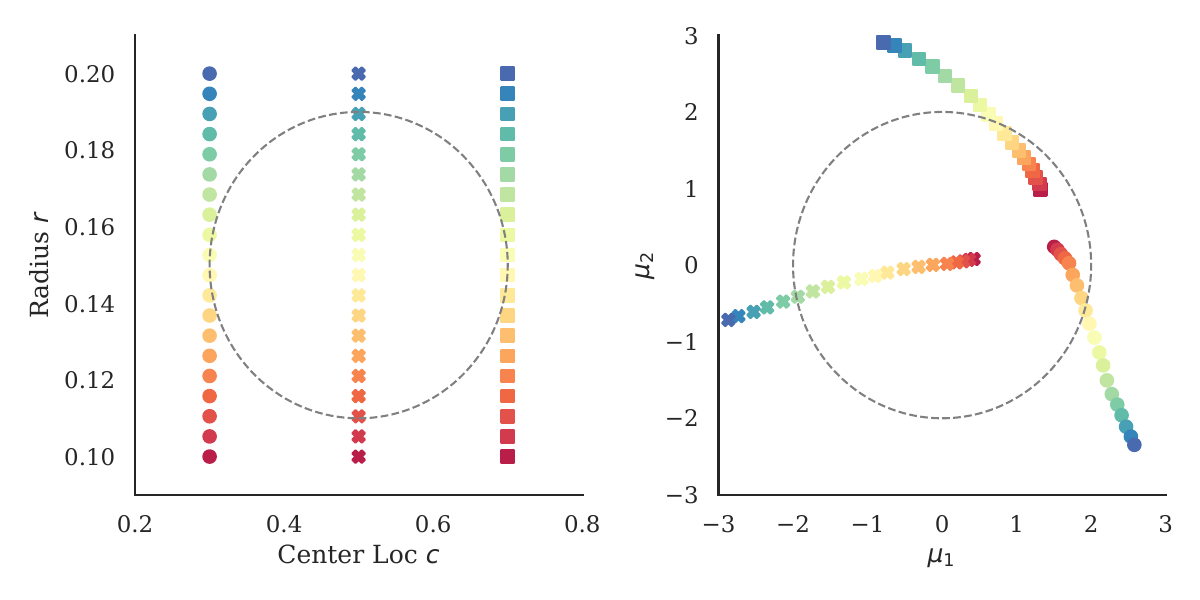}
			\par\end{centering}
	}\caption{Figure depicting the discrepancy between the true and predicted latent representations of the encoder of a VAE with two-dimensional latent code trained with in-control samples. For each subfigure, plots on the left show where real factors of variation are sampled from and the figure on the right is what the VAE encoder infers as the mean of the proposal distribution. In all figures, the regions that are considered to be in-control are represented with a dashed circle.\textbf{Top:} Real factors of variation are generated at three fixed levels of radius $r$ and varying values of center location $c$ on the left figure. Corresponding inferred means are plotted on the right graph. \textbf{Bottom:} Similar to (b) but the center location $c$ fixed at three levels and varying $r$.}
	\label{fig:proposals}
\end{figure}

In this section, we will investigate the latent representations produced by the encoder and whether it can be mapped back to the "true" latent space that generates the data in the context of our simulation study. 

We first train a VAE with an architecture described in \ref{tab:model-architectures} and fix the generating latent representation as $ r=2 $.
The training samples are generated by the in-control dome generation process as described in \ref{sec:simsetting}. 
We will use the encoder of the trained VAE for the rest of the analysis.

We can generate samples from the trained encoder by fixing one of the true latent factors and traversing along the other. 
The plots on the left side of \ref{fig:proposals} depict the traversals of the true latent space we sample the domes from. 
We then push these generated examples through the encoder to obtain their respective proposal distributions. We will compare the mean of the respective proposal distributions and the true latent space. If the learned proposal distribution is mapped into a substantially different geometry by the encoders, we will describe the distribution as "incorrect". 
 
\ref{fig:proposals} shows the incorrectness in the mapping of latent representations. This incorrect mapping behavior is even worse when we are dealing with the extreme values in the true latent space. 
For example, from \ref{fig:proposals} (b), we can conclude that domes with extremely small radii will likely go undetected if only the latent-space statistic is used. 

Overall, the learned latent representations are typically "incorrect" especially for the samples with extreme latent variables. 
This, in turn, will lead to an incorrect out-of-control assignment in Phase-II analysis, if only the latent-space monitoring statistic is used.

\subsection{On the Extrapolation Performance of the Decoder \label{sec:simstudy:generator}}

\begin{figure}[t]
\includegraphics[width=0.9\linewidth]{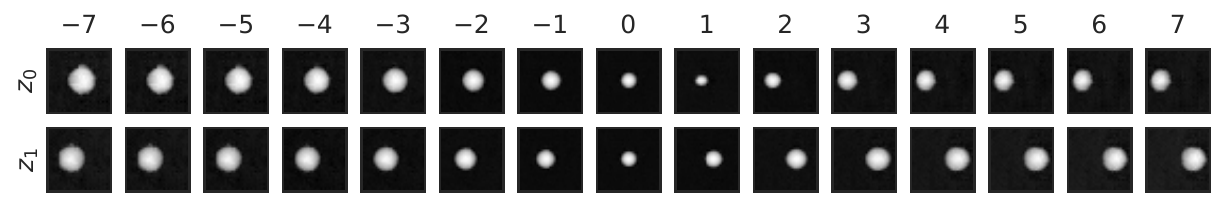} \caption{latent-space traversal and the response of the decoder of a VAE with
2-dimensional latent codes and trained with in-control dome samples.
Each row represents which latent dimension is traversed while the
other dimension is fixed at zero. Each column represents what value
is assigned to that latent dimension that is represented by the row
label. Each image in each cell is generated by the decoder using that
specific latent variable combination.}
\label{fig:manifold_vae}
\end{figure}

In this subsection, we will evaluate the extrapolation performance of the decoder. To demonstrate this, we showed the generated images by the decoder in \ref{fig:manifold_vae}, when traveling along one axis of the latent dimension while keeping the other fixed. 

Here, the decoder is trained on in-control samples described in \ref{sec:simsetting}, which is the same VAE described in \ref{sec:simstudy:recognition}

It should be cross-examined with \ref{fig:proposals} above as the encoder and decoder are tightly coupled to each other. 
We observe two important behaviors: the posterior gets distorted beyond two or
three standard deviations, and the representations are partially entangled
in line with the behavior of its encoder depicted in \ref{fig:proposals}.

To see how this will help to detect disturbances in the latent space, we consider a dome that is extremely small in terms of the radius (i.e., small $r$) or at the very margins of the grid in terms of center location (i.e., center location $c$ far from 0.5). 
Looking at \ref{fig:manifold_vae}, we can observe that the decoder simply cannot generate such a sample because it does not extrapolate well in either of the latent dimensions. This will, in turn, produce a larger reconstruction error and can be captured by the residual-space monitoring statistic.

Recall once again that the disturbance described is purely on the latent distribution $p(\mbz)$ and yet detected by the residual-space monitoring statistic only due to the extrapolation issue. 

\subsection{On the Estimation of Log-likelihood Under Importance Sampling}

Earlier, we claimed that it would take too many Monte Carlo
iterations to get a meaningful estimate of ERE defined as $\E_{\mbz\sim q_{\mbtheta}}\log\decoding$.
In this section, we test that claim on a random in-control sample
$\mbx$ using the proposal distribution $\mbz\sim\encoding$, which
is obtained via the encoder of the same VAE model we have been using in this section. The results of the sampling-based estimation of ERE,
first-order approximation $ERE_{1}$, and second-order approximation
$ERE_{2}$ are shown in \ref{fig:Estimation-comparison-between} below.
The key observation is that it takes at least 60 Monte
Carlo iterations to get a stable and accurate estimation. At that
level, the single pass through the encoder is negligible. This means
using sampling will be more costly at least 60 samples to achieve
the same accuracy as the first-order approximation that we suggest
and at least 80 samples to get the accuracy of the second-order approximation.
Another important observation is that the second-order approximation is
a bit more accurate than first-order approximation since it is closer
to the sample-average approximation, but their difference is quite insignificant.
Furthermore, it requires much more computation for the second-order approximation, given the second-order Hessian matrix needs to be evaluated. In the next subsection, we
will evaluate the performance of $ERE_{2}$ and $ERE_{1}$ in Phase-II
monitoring to evaluate whether the added computational complexity
for $ERE_{2}$ is justifiable.

\begin{figure}[t]
\includegraphics[width=0.9\textwidth]{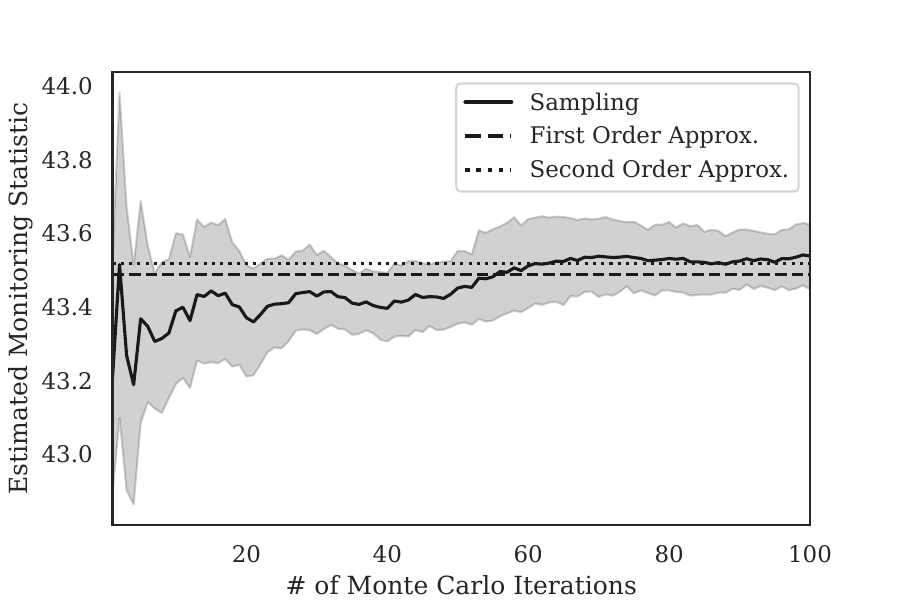}

\caption{Estimation comparison between Monte Carlo sampling, first-order approximation
and second-order approximation. 95\% confidence interval band is shown
in the gray band and is based on simulations with ten different seeds.
\label{fig:Estimation-comparison-between}}
\end{figure}

\subsection{Comparison of Detection Performance of Proposed Statistics}

We now compare the proposed statistics based on the Phase II monitoring performance by how accurately they
detect profiles from out-of-control processes outlined in \ref{sec:simsetting}.
Note that for all statistics that require sampling, we obtain a single
sample and calculate the statistic based on that to keep the computational
demand the same for all statistics and emulate the computational constraints
of a real-life case. A preliminary result we must check is the robustness
of the statistics by making sure all proposed statistics have false
alarm rates on the held-out in-control test set, which should also
be less than the desired rate 5\%. \ref{tab:far} demonstrates that
this is the case for all of them.

\begin{table}[t]
\global\long\def\arraystretch{1.3}%
 \caption{False alarm rates on the held-out dataset averaged over 10 replications
per model and monitoring statistic. Standard deviations are in parentheses.\label{tab:far}}

\begin{tabular}{llllll}
\toprule 
Statistic & $ERE_{1}$ & SPE/R & D & $H^2$ & $T^2$\tabularnewline
 & 0.041(0.006) & 0.051(0.005) & 0.044(0.004) & 0.052(0.005) & 0.043(0.009)\tabularnewline
\bottomrule
\end{tabular}
\end{table}

\begin{figure}[!t]
\includegraphics[width=1\linewidth]{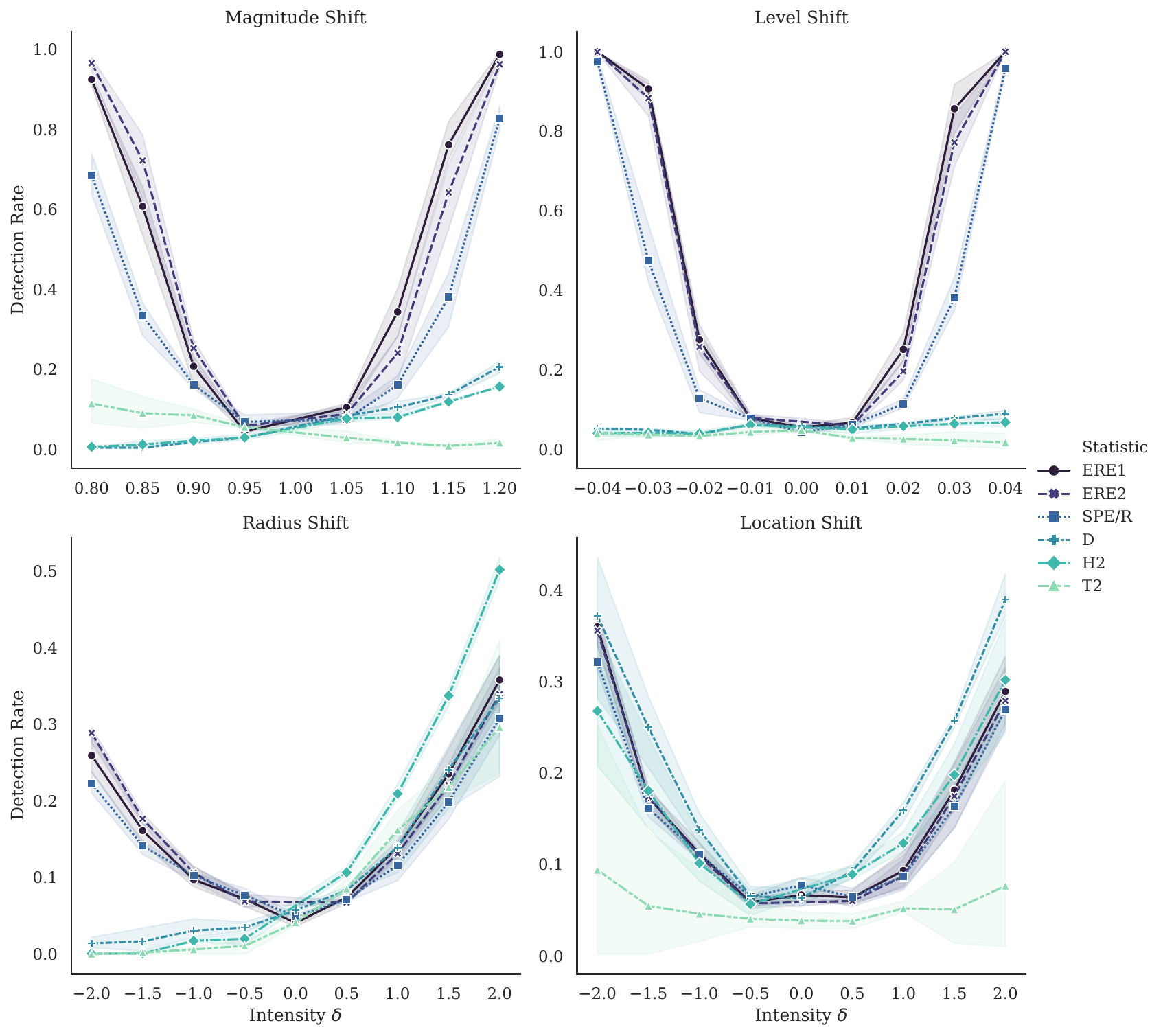}
\caption{Fault detection rates (y-axis) for varying intensities (x-axis) of
different disturbance types (quadrants). Bands represent a 95\% confidence
interval estimated around mean detection rates.}
\label{fig:disturbance_on_pxz}
\end{figure}

Through \ref{fig:disturbance_on_pxz}, we observe a clear superiority
of $ERE_{1}$ and $ERE_{2}$ over other methods when the disturbance
is on the observable space (top row). latent-space statistics
$D$, $H^{2}$ and $T^{2}$ fail in this case since that they are
purely computed using the proposal distribution latent variables.
$ERE_{1}$ and $ERE_{2}$ also outperform $SPE/R$, although by a
smaller margin it has with the latent variable-statistics. Between
$ERE_{1}$ and $ERE_{2}$, it's hard to claim which one works better
since their mean performances are quite close to each other.

For the latter two disturbances occurring purely on latent dimensions,
results are presented in the bottom row of \ref{fig:disturbance_on_pxz}.
The key observations can be listed as follows:
\begin{itemize}
\item Generally $ERE_{1}$ and $ERE_{2}$,
$D$ and $H^{2}$ tend to perform better than $SPE/R$ and $T^{2}$.
A commonality between the former three is that they do not rely on
random samples, supporting our argument against this practice.
\item Observe the radius shift-type disturbance show in the bottom left
figure. Even though $H^{2}$ performs better on positive intensities
(larger radii), it completely misses negative intensities (smaller
radii). We foresaw this result in \ref{sec:simstudy:recognition}.
To reiterate, the "incorrect" mapping of the latent space and the lack of extrapolation in the encoder is the reason behind this. We would also suggest that this result can extend to all the latent-variable based statistics for deep autoencoder-based methods.
\item Unlike latent-space statistics, $ERE_{1}$ and $ERE_{2}$
and $SPE/R$ behave more robustly against varying intensities. In
other words, the detection rate increase with increased intensities
consistently. Among these, we observe that $ERE_{1}$ and $ERE_{2}$
consistently outperform $SPE/R$.
\item $ERE_{1}$ and $ERE_{2}$ perform very similarly. In this case, we
conclude that the second-order information does not help too much
for Phase-II monitoring. The reason behind this is that the second-order
information also comes from the encoder. However, given that the encoders are trained on in-control samples and may provide inaccurate information in the out-of-control regions, the second-order information for out-of-control samples would be biased. 
Therefore, it does not provide additional gain for monitoring performance.
\end{itemize}

As mentioned, in a real-life process, disturbances on the residual
space is often more likely than the disturbance in the latent space.
Therefore, we would recommend the use of residual-space monitoring
statistics. Among all residual-space monitoring statistics, we conclude
that $ERE_{1}$ perform the best, considering the accuracy, robustness,
and computational demand. This will be further validated through the
case study analysis.

\section{Case Study Analysis \& Results}

\label{sec:case-study} 
In this section, we will evaluate the performance of the proposed algorithm using a real case study. 
Our dataset consists of defect image profiles from a hot-steel rolling
process, which is shown in \ref{fig:Rolling}. There are 13 classes
of surface defect types identified by the domain engineers. Four of
these classes---0,1,9 and 11---are considered minor defects and
they constitute our in-control set. There are 338 images
in these classes. The other nine classes make up the out-of-control
cases and they have in combination 3351 images to report detection
accuracy for. We randomly partition the in-control corpus to fix train,
validation, and test sets with 60\%-20\%-20\% relative sizes, respectively.
The rest of the procedure followed is outlined in \ref{sec:methodology:procedure}.
Same as in the simulation study, to account for randomness in weight
initialization, we replicate the experiment with 10 different seeds.
For comparison, we also include the monitoring performance with the
traditional PCA method with the same residual-space control chart,
denoted as PCA-Q. The results are summarized in \ref{tab:rolling_results}
below.

\begin{sidewaystable}
\global\long\def\arraystretch{1.3}%
 \caption{Summary of fault detection rates on out-of-control cases averaged
over 10 replications per model and monitoring statistic. Standard
deviations are in parentheses. Bolded values represent the maximum
average across different statistics. \label{tab:rolling_results}}

\centering{}%
\begin{tabular}{llllllll}
Model & \multicolumn{6}{c}{VAE} & \multicolumn{1}{c}{PCA}\tabularnewline
Statistic & D & $H^2$ & $T^2$ & SPE/R & ERE & ERE2 & Q\tabularnewline
Fault ID &  &  &  &  &  &  & \tabularnewline
\midrule 
2 & 0.00(0.00) & 0.00(0.00) & 0.00(0.00) & 0.37(0.03) & 0.44(0.06) & \textbf{0.50}(0.06) & 0.00(0.00)\tabularnewline
3 & 0.17(0.06) & 0.23(0.04) & 0.03(0.03) & 0.84(0.01) & 0.85(0.01) & \textbf{0.86}(0.01) & 0.78(0.00)\tabularnewline
4 & 0.00(0.00) & 0.00(0.00) & 0.00(0.00) & 0.62(0.02) & \textbf{0.75}(0.05) & 0.71(0.05) & 0.56(0.00)\tabularnewline
5 & 0.58(0.07) & 0.62(0.09) & 0.00(0.00) & \textbf{1.00}(0.00) & \textbf{1.00}(0.00) & \textbf{1.00}(0.00) & 0.99(0.00)\tabularnewline
6 & 0.06(0.03) & 0.15(0.08) & 0.05(0.05) & 0.79(0.01) & \textbf{0.80}(0.01) & \textbf{0.80}(0.00) & 0.52(0.00)\tabularnewline
7 & 0.01(0.01) & 0.01(0.01) & 0.00(0.00) & 0.13(0.01) & \textbf{0.17}(0.01) & 0.15(0.00) & 0.11(0.00)\tabularnewline
8 & 0.00(0.00) & 0.00(0.00) & 0.00(0.00) & 0.64(0.02) & \textbf{0.70}(0.07) & 0.69(0.01) & 0.34(0.00)\tabularnewline
10 & 0.00(0.00) & 0.00(0.00) & 0.00(0.00) & 0.49(0.03) & \textbf{0.57}(0.05) & \textbf{0.57}(0.04) & 0.29(0.00)\tabularnewline
12 & 0.00(0.00) & 0.00(0.00) & 0.00(0.00) & 0.79(0.01) & \textbf{0.80}(0.02) & \textbf{0.80}(0.02) & 0.69(0.00)\tabularnewline
13 & 0.00(0.00) & 0.00(0.00) & 0.01(0.00) & 0.71(0.04) & \textbf{0.77}(0.02) & 0.76(0.02) & 0.56(0.00)\tabularnewline
\bottomrule
\end{tabular}
\end{sidewaystable}

From \ref{tab:rolling_results}, we can observe that $ERE_{1}$ and
$ERE_{2}$ consistently outperforms all other monitoring statistic
formulations. The divide between residual-space statistics and latent-space statistics observed in the simulation study is further
validated here too. The inferiority of latent-space statistics
is much more obvious here in the real case study, as we observe for
most out-of-control classes, the detection rate is simply zero. This
observation further validates our claims that in practice, for deep
autoencoders, the change happens in the residual space rather than
the latent space. The advantage of VAE over PCA is mainly due to the
better representative power and data compression ability of deep autoencoders
compared to PCA. It is worth noting that the superiority of VAE over
PCA for process monitoring was also demonstrated in the earlier works
in various applications \parencite{Zhang2019-lu,wang2019systematic,lee2019process}.

\begin{figure}[t]
\begin{minipage}[t]{0.25\textwidth}%
\subfloat[Original profiles\label{fig:Original-profiles}]{\includegraphics[width=0.99\textwidth]{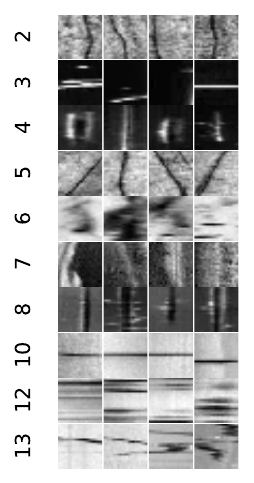}}%
\end{minipage}\hfill{}%
\begin{minipage}[t]{0.25\textwidth}%
\subfloat[Reconstructions via VAE\label{fig:Reconstructions-via-VAE}]{\includegraphics[width=0.82\columnwidth]{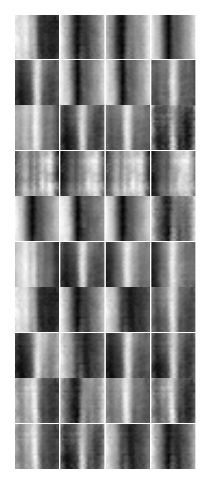}
}%
\end{minipage}\hfill{}%
\begin{minipage}[t]{0.45\textwidth}%
\subfloat[Inferred means\label{fig:Inferred-means}]{\includegraphics[width=1\textwidth]{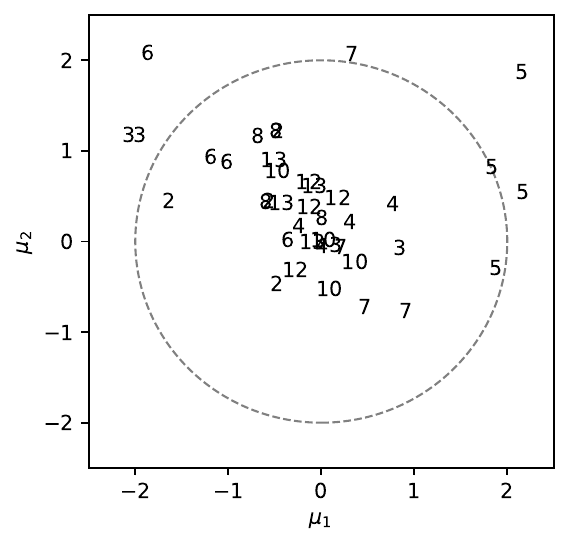}

}%
\end{minipage}\caption{Output of the VAE decoder and the encoder for randomly select rolling
profiles. \textbf{Left: }Original profiles visualized. Each row is
a class of defect profile and each column is randomly selected from
that class. \textbf{Middle: }Reconstructions of the samples with one-to-one
correspondence to the samples on the image to the left. \textbf{Right:
}Inferred mean locations of each of the defects visualized on the left.
Points are annotated by their class IDs. \label{fig:Output-of-the}}
\end{figure}

To support our claim of the ineffectiveness of latent-space
statistics, we refer the reader to \ref{fig:kde} below. We observe
how well separated the statistics are for $ERE_{1}$ and $SPE/R$
while for latent-space statistics, the obtained values are
mostly overlapping. Note that we omitted $ERE_{2}$ because it was
almost identical to $ERE_{1}$. To obtain a deeper understanding of
the results, we point out in \ref{fig:Output-of-the} for the original
images and their reconstructions. The decoder is persistent on generating
samples that look like in-control rolling samples with little fidelity
to how the original defect sample looks like. When \ref{fig:Original-profiles}
and \ref{fig:Reconstructions-via-VAE} are cross-examined, it is apparent
why reconstruction error would be high. On the contrary, \ref{fig:Inferred-means}
shows that most latent representations fall into the region that would
be considered in-control from a profile monitoring perspective. We
observed instances of classes 3,5,6 and 7 generate the latent variables
in the out-of-control regions. However, even for these classes, $SPE/R$,
$ERE_{1}$ and $ERE_{2}$ yields much better detection power than
$D$, $H^{2}$, and $T^{2}$, as it can be seen in \ref{tab:rolling_results}.
In conclusion, we would like to suggest the use of $ERE_{1}$ for
deep autoencoders, which is consistent with our findings in the simulation
study.

\begin{figure}[t]
\begin{subfigure}[b]{0.3\textwidth}
     \centering
     \includegraphics[width=\textwidth]{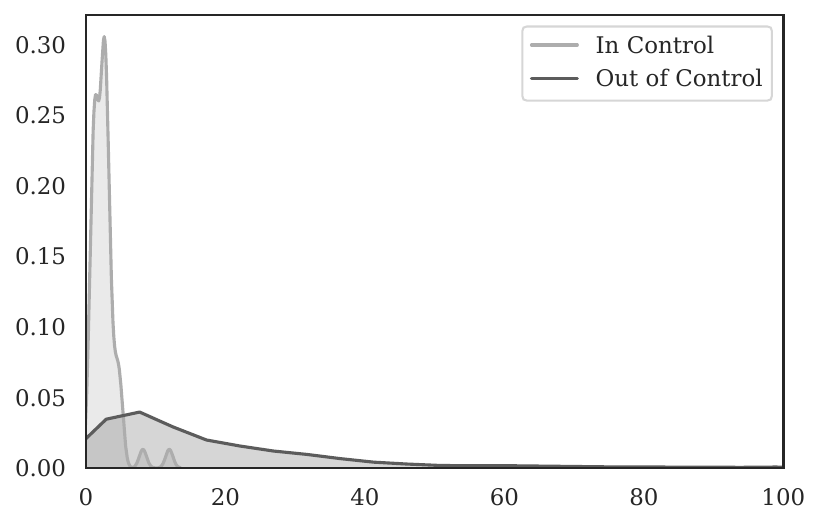}
     \caption{$ERE_{1}$}
     \label{fig:kde:ERE}
 \end{subfigure}
 \hspace{0.1\textwidth}
 \begin{subfigure}[b]{0.3\textwidth}
     \centering
     \includegraphics[width=\textwidth]{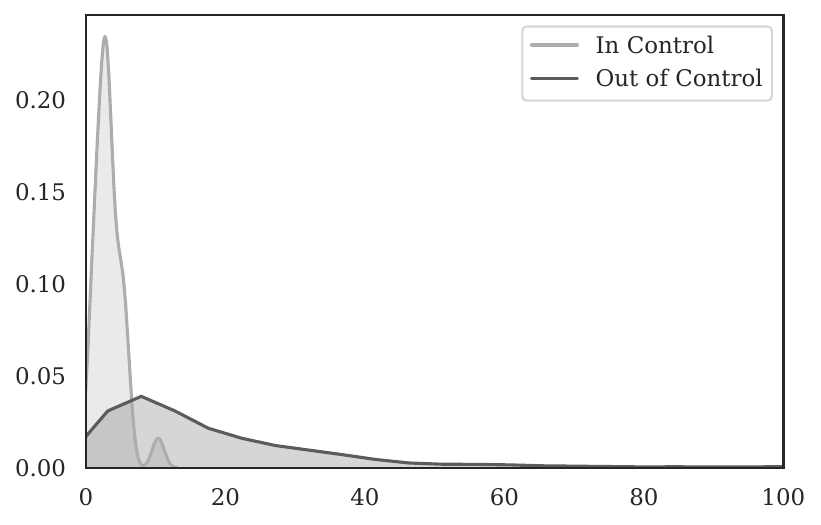}
     \caption{$SPE/R$}
     \label{fig:kde:spe-r}
 \end{subfigure}
 
\begin{subfigure}[b]{0.3\textwidth}
     \centering
     \includegraphics[width=\textwidth]{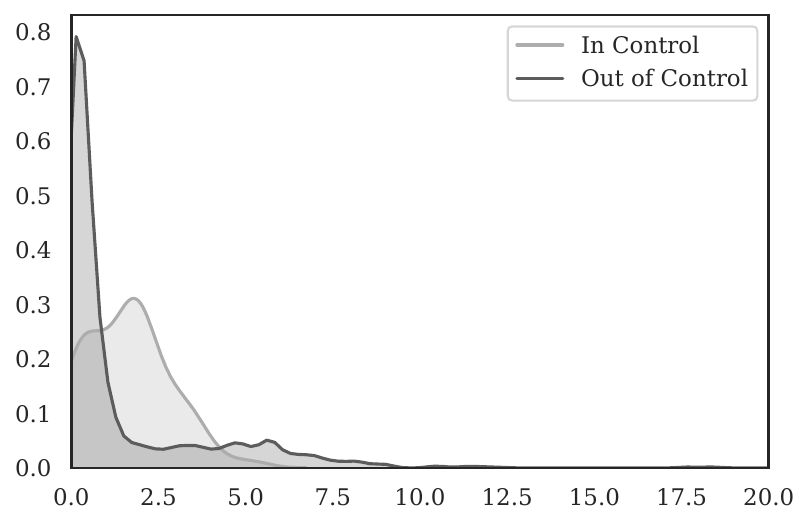}
     \caption{$H^{2}$}
     \label{fig:kde:h2}
\end{subfigure}
\hfill
\begin{subfigure}[b]{0.3\textwidth}
     \centering
     \includegraphics[width=\textwidth]{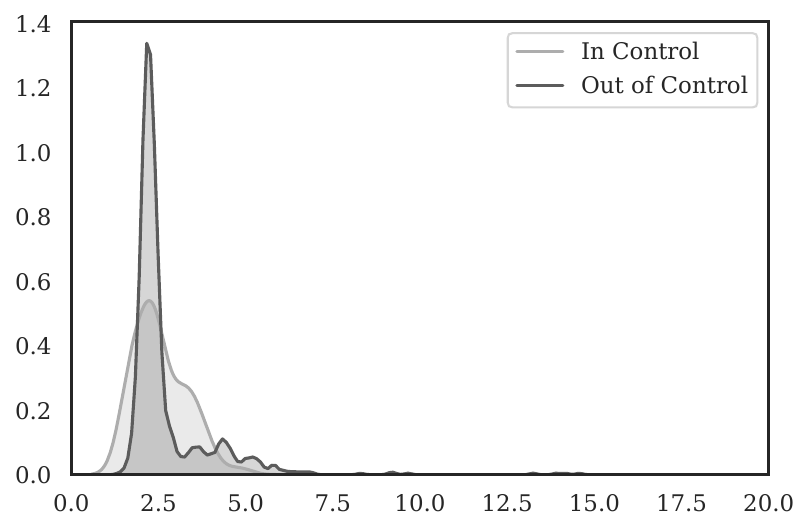}
     \caption{$D$}
     \label{fig:kde:D}
\end{subfigure}
\hfill
\begin{subfigure}[b]{0.3\textwidth}
     \centering
     \includegraphics[width=\textwidth]{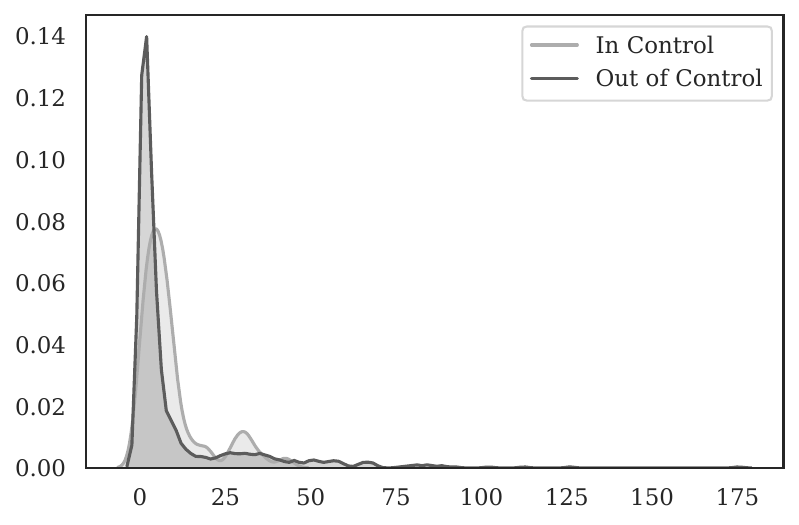}
     \caption{$T^{2}$}
     \label{fig:kde:T2}
\end{subfigure}

\caption{Kernel density estimation plots of statistics obtained for in-control
and out-of-control steel defect profiles, per each proposed statistic
type.\label{fig:kde}}
\end{figure}

Finally, we report execution time details for our proposed statistic, $ ERE_{1}] $. For this study, we utilized a workstation with 6-core Intel(R) Core(TM) i7-5930K CPU \@ 3.50GHz CPUs and 4 GeForce GTX 1080 Ti GPUs. Neural network computations are executed on a single GPU and a single CPU core is used for image input/output and preprocessing steps such as resizing to 64-by-64 and grayscale conversion wherever needed. A single GPU has 12GB memory and the model parameters take up about 730MBs. GPUs can leverage parallel computation of multiple images, therefore the remaining memory can be used to stock up images so their execution becomes parallel. An example of a batch of 128 images takes up only 63MBs more space in the GPU's memory and the per image execution time is roughly 0.8 milliseconds. On the extreme case of using a single image per batch, per image execution time is around 2 milliseconds on average, which satisfies the real-time monitoring constraint.

\section{Conclusion \label{sec:conclusions}}

In this paper, we focused on evaluating Phase-II monitoring statistics
proposed so far in the literature for VAE and demonstrate that they
were not performing optimally in terms of accuracy and/or computational
feasibility. 
First, we classified these statistics into two groups and showed how they are designed as an extension to the classical statistics used for PCA.
Then we pointed out that such an extension is not as straightforward as it seems due to the incorrectness of learned latent representations by VAEs and also due to the failure to extrapolate behavior. 
This led us to the conclusion that only residual-space statistics should be monitoring, regardless of the anticipated source of the shift in the process.
We also pointed out that the residual-space statistics based on sampling will require
too many samples to be computationally feasible. 
Finally, we proposed a novel formulation by deriving the Taylor expansion of the expected reconstruction error that addresses the computational efficiency issue in residual-space statistics.

We put our claims to the test with a carefully designed simulation study.
This study demonstrated the discrepancy between the true latent variations and its learned counterparts, and its implications to the process monitoring performance of latent-space statistics.
We also reinforced our claim that the derived statistics based on the residual space is overall more robust and accurate than all the other statistics proposed so far. 
Finally, we validated the superiority of our formulation on a real-life case study, where steel defect image profiles are used.

For future work, we hope to extend the proposed method for other types of data format. For example, for sequential profiles (e.g., time series), one-dimensional convolutional layers or a recurrent neural network for encoder and decoder structures as outlined in \textcite{ChungKDGCB15} can be used. We are also curious to see how new developments in deep learning research will affect profile monitoring in high dimensions in the future. Specifically, developments in deep latent variable models and representation learning may have important implications.

\printbibliography

\appendix

\section{Proof of Proposition 3.1 \label{sec:PoofOfPropTQ}}

The Kullback-Leibler divergence between two multivariate Gaussian
distributions has a closed-form solution. If we define these distributions
as $p_{0}=N(\mbz;\mbmu_{0},\mbSigma_{0})$ and $p_{1}=N(\mbz;\mbmu_{1},\mbSigma_{1})$
where $\mbmu$ and $\mbSigma$ are respective mean vectors and covariance
matrices, then according to \textcite{hershey2007approximating}
the closed-form solution will be the following: 
\begin{align}
\KL{p_{0}}{p_{1}} & =\frac{1}{2}[\log\frac{\g\mbSigma_{1}\g}{\g\mbSigma_{0}\g}+Tr(\mbSigma_{1}\inv\mbSigma_{0})-r+(\mbmu_{0}-\mbmu_{1})^{\top}\mbSigma_{1}\inv(\mbmu_{0}-\mbmu_{1})]\label{eq:kld-closed-form}
\end{align}
Since $\encoding=\Norm(\mbmu(\mbx),\mbSigma_{z})$ and $p(\mbz)=\Norm(0,\mbI)$,
we can derive that 
\begin{align}
\KL{\encoding}{p(\mbz)} & =\frac{1}{2}\left[-\log\g\mbSigma_{z}\g+Tr(\mbSigma_{z})-r\right]+\frac{1}{2}\mbmu(\mbx)^{\top}\mbmu(\mbx)\nonumber \\
 & =\frac{1}{2}\mbmu(\mbx)^{\top}\mbmu(\mbx)+C,\label{eq:kld-prior}
\end{align}
where $C=-\log\g\mbSigma_{z}\g+Tr(\mbSigma_{z})-r$ is a constant,
which doesn't depend on $\mbx$.

To derive the SPE statistics, we will derive

\begin{align}
 & \mathbb{E}_{\mbz\sim q_{\mbtheta}}\|\mbx-\mbW\mbz\|^{2}\nonumber \\
= & \mathbb{E}_{\mbz\sim q_{\mbtheta}}(\mbx^{\top}\mbx-2\mbz^{\top}\mbW\mbx+\mbz^{\top}\mbW^{\top}\mbW\mbz)\nonumber \\
= & \mbx^{\top}\mbx-2\mbmu(\mbx)^{\top}\mbW\mbx+\mathbb{E}_{\mbz\sim q_{\mbtheta}}(\mbz^{\top}\mbW^{\top}\mbW\mbz)\label{eq: spew}
\end{align}

Here, we know that 
\begin{align}
 & \mathbb{E}_{\mbz\sim q_{\mbtheta}}(\mbz^{\top}\mbW^{\top}\mbW\mbz)\nonumber \\
= & \mathbb{E}_{\mbz\sim q_{\mbtheta}}tr(\mbz^{\top}\mbW^{\top}\mbW\mbz)\nonumber \\
= & tr\left(\mbW^{\top}\mbW\mathbb{E}_{\mbz\sim q_{\mbtheta}}(\mbz\mbz^{\top})\right)\nonumber \\
= & tr\left(\mbW^{\top}\mbW(\mbmu(\mbx)\mbmu(\mbx)^{\top}+\Sigma_{z})\right)\nonumber \\
= & \mbmu(\mbx)^{\top}\mbW^{\top}\mbW\mbmu(\mbx)+tr\left(\mbW^{\top}\mbW\Sigma_{z}\right)\label{eq: tracezwwz}
\end{align}

Therefore, by plugging \ref{eq: tracezwwz} into \ref{eq: spew},
we have 
\begin{align}
\mathbb{E}_{\mbz\sim q_{\mbtheta}}\|\mbx-\mbW\mbz\|^{2} & =\mbx^{\top}\mbx-2\mbmu(\mbx)^{\top}\mbW\mbx+\mathbb{E}_{\mbz\sim q_{\mbtheta}}(\mbz^{\top}\mbW^{\top}\mbW\mbz)\nonumber \\
 & =\mbx^{\top}\mbx-2\mbmu(\mbx)^{\top}\mbW\mbx+\mbmu(\mbx)^{\top}\mbW^{\top}\mbW\mbmu(\mbx)+tr\left(\mbW^{\top}\mbW\Sigma_{z}\right)\nonumber \\
 & =\|\mbx-\mbW\mbmu(\mbx)\|^{2}+C\label{eq:q-to-ere}
\end{align}
where $C=tr\left(\mbW^{\top}\mbW\Sigma_{z}\right)$ that does not
depend on $\mbx$.

\section{A Toy Example to Demonstrate Out-of-distribution Behavior of Neural
Networks \label{app:rosenbrock}}

\begin{figure}[t]
\begin{centering}
\includegraphics[width=1\textwidth]{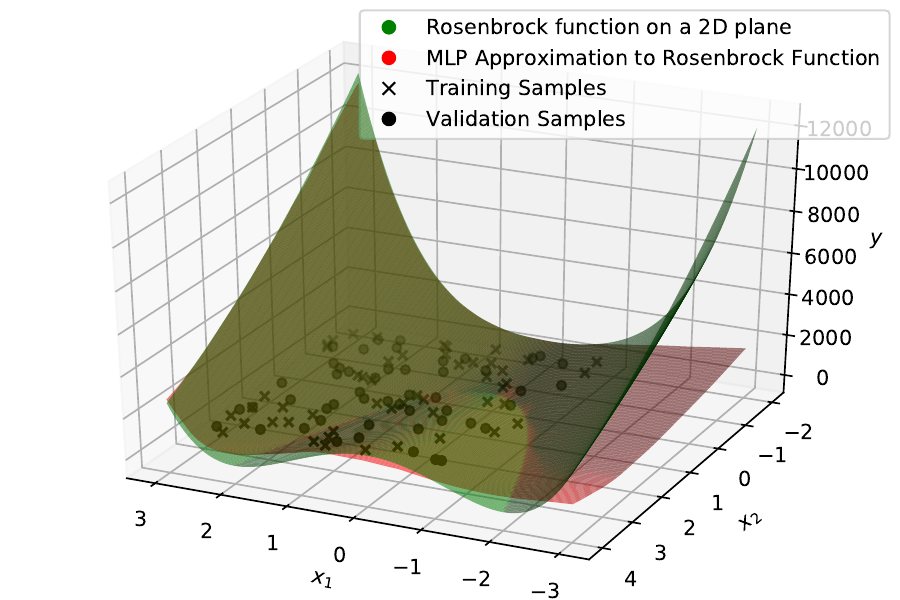}
\par\end{centering}
\caption{Rosenbrock function (green surface) approximated by an multilayer
perceptron(red surface) given training (black crosses) and validation
(black dots) samples form a bounded region $(x_{i},y_{i})\in[-1,3]\times[-2,3]$.\label{fig:Rosenbrock}}
\end{figure}

Assume using a multilayer perceptron, we are trying to approximate
the famous Rosenbrock function $f(x,y)=(a-x)^{2}+b(y-x^{2})^{2}$
given $(a,b)=(1,100)$. In this small experiment, we sample tuples
of two-dimensional points from a bounded region $(x_{i},y_{i})\in[-1,3]\times[-2,3]$.
We use a multilayer perceptron with six hidden layers and a hundred
neurons in each layer. Half of the points are used in training, and
the other half is used as a validation set to optimize hyper-parameters.
Using the trained network, we plot the actual Rosenbrock function
along with the neural network approximation in \ref{fig:Rosenbrock}.
Notice how well the function is approximated for the region $[-1,3]\times[-2,3]$,
but there is a serious discrepancy between the approximated and the
real outside of the region. This is a small yet to the point example
of out-of-distribution issues with neural networks.

\section{ERE Testing Statistic Derivation \label{app:ere}}

To derive the $ERE_{1}$ and $ERE_{2}$, we first define $R(z)=\|y-\mbmu_{\mbtheta}(z)\|^{2}$
as the reconstruction error (RE). The quantity we would like approximate
is $E_{\mbz\sim q_{\phi}}R(\mbz)$ where $\encoding=\Norm(\mbmu_{\mbphi}(\mbx),\Sigma_{z})$.
We are looking for the Taylor expansion of the expected RE (ERE) around
$z_{0}=\mbmu_{\mbphi}(\mbx)$, i.e., the first moment. For notational
simplicity, we use $H_{z}$ to denote the Hessian $R''(\mbmu_{\mbphi}(\mbx))$.
The derivation is formalized as follows:

\begin{align}
E_{\mbz\sim q_{\phi}}R(\mbz) & =R(\mbmu_{\mbphi}(\mbx))+R'(\mbmu_{\mbphi}(\mbx))E_{\mbz\sim q_{\phi}}[\mbz-\mbmu_{\mbphi}(\mbx)])\nonumber \\
 & \quad+\frac{1}{2}E_{\mbz\sim q_{\phi}}[(\mbz-\mbmu_{\mbphi}(\mbx))^{\top}\mathbf{H}_{z}(\mbz-\mbmu_{\mbphi}(\mbx))]+O(\|(\mbz-\mbmu_{\mbphi}(\mbx)\|^{3}\nonumber \\
 & \simeq R(\mbmu_{\mbphi}(\mbx))+\frac{1}{2}E_{\mbz\sim q_{\phi}}[(\mbz-\mbmu_{\mbphi}(\mbx))^{\top}\mathbf{H}_{z}(\mbz-\mbmu_{\mbphi}(\mbx))]\nonumber \\
 & =R(\mbmu_{\mbphi}(\mbx))+\frac{1}{2}\mathrm{tr}(\mathbf{H}_{z}E[(\mbz-\mbmu_{z})(\mbz-\mbmu_{z})^{T}])\nonumber \\
 & =R(\mbmu_{\mbphi}(\mbx))+\frac{1}{2}\mathrm{tr}(\mathbf{H}_{z}\Sigma_{z})\label{eq:derived-ere2}
\end{align}

Note for $ERE_{1}$, the second term $\frac{1}{2}\mathrm{tr}(\mathbf{H}_{z}\Sigma_{z})$
is droped and we are left with $R(\mbmu_{\mbphi}(\mbx))$ only. For $ERE_{2}$,
since $\Sigma_{z}$ is a diagonal matrix, $\mathrm{tr}(\mathbf{H}_{z}S_{z})=\mathrm{tr}(diag(\mathbf{H}_{z})S_{z})=\sum_{i}(\mathbf{H}_{z})_{ii}(S_{z})_{ii}$
holds. We can utilize this result to compute $ERE_{2}$, in a more computationally
efficient manner.
\end{document}